\documentclass[10pt,twocolumn,letterpaper]{article}

\usepackage{iccv}
\usepackage{times}
\usepackage{epsfig}
\usepackage{graphicx}
\usepackage{amsmath}
\usepackage{amssymb}
\usepackage{comment}

\usepackage{balance}
\usepackage{color}
\usepackage{array}
\usepackage[toc,page]{appendix}
\usepackage{bbm}
\usepackage{algorithm}
\usepackage[noend]{algpseudocode} 
\usepackage{bbding}
\usepackage[noadjust,nosort]{cite}
\usepackage{enumitem}
\usepackage{booktabs}
\usepackage[normalem]{ulem}
\usepackage{setspace}
\usepackage{stfloats}
\usepackage{authblk}

\newcommand\thoughts[1]{}

\newcommand\rindra[1]{\textcolor{red}{#1}}

\newcommand{\bftab}{\fontseries{b}\selectfont}

\usepackage[pagebackref=true,breaklinks=true,letterpaper=true,colorlinks,bookmarks=false]{hyperref}

\iccvfinalcopy 

\def\iccvPaperID{8083} 

\ificcvfinal\fi

\begin{document}

\title{SimROD: A Simple Adaptation Method for Robust Object Detection}

\makeatletter
\newcommand\email[2][]%
   {\newaffiltrue\let\AB@blk@and\AB@pand
      \if\relax#1\relax\def\AB@note{\AB@thenote}\else\def\AB@note{\relax}%
        \setcounter{Maxaffil}{0}\fi
      \begingroup
        \let\protect\@unexpandable@protect
        \def\thanks{\protect\thanks}\def\footnote{\protect\footnote}%
        \@temptokena=\expandafter{\AB@authors}%
        {\def\\{\protect\\\protect\Affilfont}\xdef\AB@temp{#2}}%
         \xdef\AB@authors{\the\@temptokena\AB@las\AB@au@str
         \protect\\[\affilsep]\protect\Affilfont\AB@temp}%
         \gdef\AB@las{}\gdef\AB@au@str{}%
        {\def\\{, \ignorespaces}\xdef\AB@temp{#2}}%
        \@temptokena=\expandafter{\AB@affillist}%
        \xdef\AB@affillist{\the\@temptokena \AB@affilsep
          \AB@affilnote{}\protect\Affilfont\AB@temp}%
      \endgroup
       \let\AB@affilsep\AB@affilsepx
}
\makeatother

\author[1]{Rindra Ramamonjison} 
\author[1]{Amin Banitalebi-Dehkordi} 
\author[2]{Xinyu Kang} 
\author[3]{Xiaolong Bai} 
\author[1]{Yong Zhang} 
\affil[1]{Huawei Technologies Canada Co., Ltd}
\affil[2]{University of British Columbia}
\affil[3]{Huawei Cloud}
\email{\small rindranirina.ramamonjison@huawei.com, amin.banitalebi@huawei.com, xinyu.kang@alumni.ubc.ca, baixiaolong1@huawei.com, yong.zhang3@huawei.com}


\maketitle
\ificcvfinal\thispagestyle{empty}\fi

\begin{abstract}
\vspace{-4pt}

This paper presents a Simple and effective unsupervised adaptation method for Robust Object Detection (SimROD). To overcome the challenging issues of domain shift and pseudo-label noise, our method integrates a novel domain-centric augmentation method, a gradual self-labeling adaptation procedure, and a teacher-guided fine-tuning mechanism. Using our method, target domain samples can be leveraged to adapt object detection models without changing the model architecture or generating synthetic data. When applied to image corruptions and high-level cross-domain adaptation benchmarks, our method outperforms prior baselines on multiple domain adaptation benchmarks. SimROD achieves new state-of-the-art on standard real-to-synthetic and cross-camera setup benchmarks. On the image corruption benchmark, models adapted with our method achieved a relative robustness improvement of 15-25\% AP50 on Pascal-C and 5-6\% AP on COCO-C and Cityscapes-C. On the cross-domain benchmark, our method outperformed the best baseline performance by up to 8\% AP50 on Comic dataset and up to 4\% on Watercolor dataset. 
   
\end{abstract}

\vspace{-8pt}
\section{Introduction}
\vspace{-5pt}
State-of-the-art object detection models have proved to be highly accurate when trained on images that have the same distribution as the test set \cite{zou2019object}. However, they can fail when deployed to new environments due to domain shifts such as weather changes (e.g. rain or fog), light conditions variations, or image corruptions (e.g. motion blur) \cite{Stylize}. Such failure is detrimental for mission-critical applications such as self-driving, security, or automated retail checkout, where domain shifts are common and inevitable. To make them succeed in applications where reliability is key, it is important to make detection models robust to domain shifts.

Many methods have been proposed to overcome domain shifts for object detection. They can be categorized as data augmentation \cite{Stylize, AugMix, DeepAugment}, domain-alignment \cite{DAF, MAF, SCDA, C2F, SWDA, Every, DAM, PDA}, domain-mapping \cite{domain-map, cross-domain, DAM, PDA}, and self-labeling techniques \cite{self-labeled, STAC, RLDA, cross-domain}. Data augmentation methods can improve the performance on some fixed set of domain shifts but fail to generalize to the ones that are not similar to the augmented samples \cite{anonymous2021, mintun2021interaction, taori2020}. Domain-aligning methods use samples from the target domain to align intermediate features of networks. These methods require non-trivial architecture changes such as gradient reversal layers, domain classifiers, or some specialized modules. On the other hand, domain-mapping methods translate labeled source images to new images that look like the unlabeled target domain images using image-to-image translation networks. Similar to the augmentation methods, they are suboptimal since the generated images do not necessarily have a high perceptual similarity to real target domain images. Finally, self-labeling is a promising approach since it leverages unlabeled training samples form the target domain. However, generating accurate pseudo-labels under domain shift is hard; and when pseudo-labels are noisy, using target domain samples for adaptation is ineffective.


In this paper, we propose a Simple adaptation method for Robust Object Detection (SimROD), to mitigate the domain shifts using domain-mixed data augmentation and teacher-guided gradual adaptation. Our simple approach has three design benefits. First, it does not require ground-truth labels of target domain data and leverage unlabeled samples. Second, our approach requires neither complicated architecture changes nor generative models for creating synthetic data \cite{cross-domain}. Third, our simple method is architecture-agnostic and is not limited to region-based detectors. The main contributions of this paper are summarized as follows: 
\begin{enumerate}[leftmargin=*]
\item We propose a simple method to improve the robustness of object detection models against domain shifts. Our method first adapts a large teacher model using a gradual adaptation approach. The adapted teacher generates accurate pseudo-labels for adapting the student model.
\vspace{-5pt}
\item We introduce an augmentation procedure called \textit{DomainMix} to help learn domain-invariant representations and reduce the pseudo-label noise that is exacerbated by the domain shift. \textit{DomainMix} efficiently mixes the labeled images from the source domain with the unlabeled samples from the target domain along with their (pseudo-)labels and gives strong supervision for self-adaptation. The mixed training samples are used for adapting both the teacher and student models.
\vspace{-6pt}
\item We conduct a comprehensive and fair benchmark to demonstrate the effectiveness of SimROD to mitigate different kinds of shifts including synthetic-to-real, cross-camera setup, real-to-artistic, and image corruptions. We show that our simple method can achieve new state-of-the-art results on some of these benchmarks. We also conduct ablation studies to provide insights on the efficiency and effectiveness of our method. 
\end{enumerate}

\vspace{-4pt}
\section{Motivation and related works} \label{related_works}
\vspace{-4pt}
In this section, we review the mainstream approaches relevant to our work and explain the motivation of our work.
\vspace{-20pt}
\subsubsection*{Data augmentations for robustness to image corruption}
\vspace{-6pt}
Data augmentation is an effective technique for improving the performance of deep learning models. Recent works have also explored the role of augmentation in enhancing the robustness to domain shifts. In particular, specialized augmentation methods have been proposed to combat the effect of image corruptions for image classification \cite{hendrycks2019benchmarking, AugMix, DeepAugment} and object detection \cite{Stylize,cygert2020toward}. For example, AugMix \cite{AugMix} samples a set of geometric and color transformations which are applied sequentially to each image and mixes the original image with multiple augmented versions. Subsequently, DeepAugment \cite{DeepAugment} generates augmented samples using image-to-image translation networks whose weights are perturbed with random distortions. \cite{Stylize,cygert2020toward} proposed style transfer \cite{Geirhos19} as data augmentation for increasing the shape bias and improve robustness to image corruptions. 

While these augmentation methods offer some improvement over the source baseline, they can overfit to few corruption types and fail to generalize to others. In fact, \cite{anonymous2021} provided empirical evidence that the perceptual similarity between the augmentation transformation and the corruption is a strong predictor of corruption error. \cite{anonymous2021} also observed that broader augmentation schemes perform better on dissimilar corruptions than more specialized ones. \cite{taori2020} showed that augmentation techniques that are tailored to synthetic corruptions have difficulty to generalize to natural distributions shifts. In their extensive study, training on more diverse data was the only intervention that effectively improved the robustness to natural distribution shifts.

\vspace{-6pt}
\subsubsection*{Unsupervised domain adaptation for object detection}
\vspace{-6pt}

Unsupervised domain adaptation (UDA) methods leverage unlabeled images from the target domain to explicitly mitigate the domain shift. In contrast to images obtained with augmentation, these unlabeled samples are more similar to the test samples as they are from the same domain. Moreover, leveraging unlabeled samples is practical since they are cheap to collect and do not require laborious annotation.

Several approaches have been proposed to solve the UDA problem for object detection. Adversarial training methods such as \cite{DAF} learn domain-invariant representations of two-stage detector networks. Recent methods improved the performance, by mining important regions and aligning at the region-level \cite{MAF}, by using hierarchical alignment module \cite{SCDA}, by coarse-to-fine feature adaptation \cite{C2F}, or by enforcing strong local alignment and weak global alignment \cite{SWDA}. \cite{Every} proposed a center-aware alignment method for anchor-free FCOS model. While alignment methods help reduce the domain shift, they require architecture changes since extra modules such as gradient reversal layers and domain classifiers must be added to the network.

Alternatively, domain-mapping methods tackle UDA by first translating source images to images that resemble the target domain samples using a conditional generative adversarial network (GAN) \cite{domain-map, cycada}. The model is then fine-tuned with the domain-mapped images and the known source labels. For object detection, \cite{DAM, PDA} combined domain transfer with adversarial training. For instance, \cite{DAM} generates a diverse set of intermediate domains between the source and target to discriminate and learn domain-invariant features. 

Batch normalization (BN) \cite{bn} layers are prevalent in most neural
networks because they can accelerate the learning, prevent overfitting
and enable deeper networks to converge \cite{bn-optim}. Recent works
have shown that adapting BN layers can improve robustness to adversarial attacks \cite{advprop} or image corruptions \cite{bn-schneider20} and reduce domain
shifts \cite{adabn, domain-specific-bn}. 

\vspace{-8pt}
\subsubsection*{Self-training for object detection adaptation}
\vspace{-6pt}

Self-training enables a model to generate its own pseudo-labels on the unlabeled target samples. Recently, \cite{STAC} has proposed the STAC framework for semi-supervised object detection with pseudo-labels. However, pseudo-labeling can degenerate the performance in the presence of domain shift as the pseudo-labels on target samples may become incorrect leading to poor supervision. Instead, our work tackles the domain shift between the original source training data and the unlabeled target training data. To reduce domain shift, \cite{MTOR} enforced region-level and graph-structures consistencies between a mean teacher model and the student model using additional regularization loss functions. Next, \cite{RLDA} proposed a method to directly mitigate the noisy pseudo-labels of Faster-RCNN detectors by modeling their proposal distribution. Unlike \cite{RLDA}, our method is agnostic to the model architecture and can also work with single-stage object detectors too. Finally, \cite{cross-domain} combined domain transfer with pseudo-labeling and is also architecture-agnostic. 

In contrast to these prior works, our proposed adaptation method is simpler because it does not generate synthetic data using GANs, does not add new loss functions and does not change the model architecture. As it will be shown in Section \ref{experiments}, our simple method is also more effective in reducing domain shifts and label noise.

\vspace{-4pt}
\section{Problem definition and proposed solution}\label{methods}
\vspace{-7pt}
In this section, we define the adaptation problem and describe our proposed solution.

\begin{figure*}
\centering\includegraphics[width=2\columnwidth]{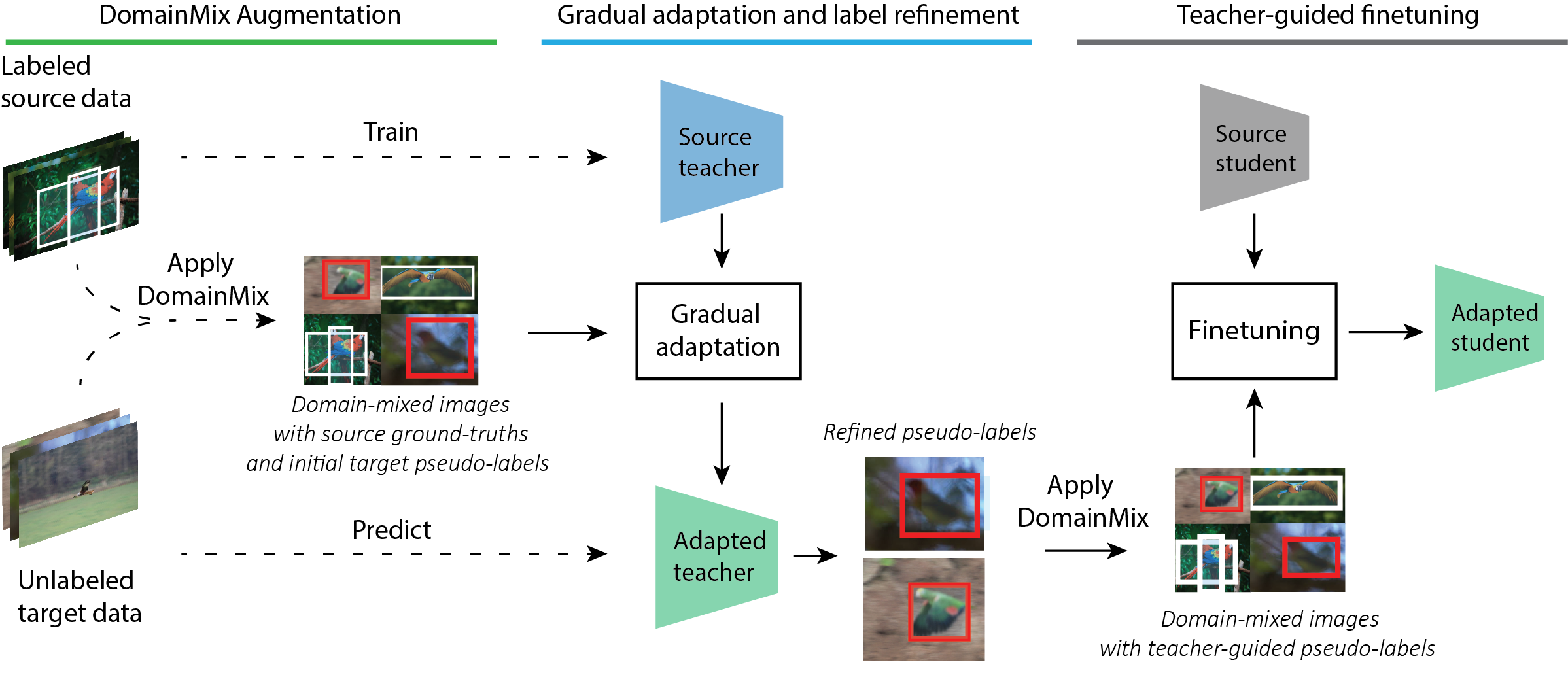}\vspace{-10pt}
\caption{\small Our proposed adaptation method for robust object detection mitigates the domain shift and label noise using three simple steps. (1) The proposed DomainMix augmentation module randomly samples and mixes images from both the source and target domains along with their ground-truth and pseudo-labels. (2) These domain-mixed images are used to gradually adapt the batch norm and convolutional layers of a large source teacher model. During this step, the pseudo-labels of the target domain images are also refined. (3) New domain-mixed images with the refined pseudo-labels are used to finetune the source student model.}
\label{sys-diagram}
\end{figure*}

\vspace{-2pt}
\subsection{Problem statement}
\vspace{-6pt}

We are given a source model $\tt M$ for an object detection task with parameters $\theta^{s}_{\tt M}$, which is
trained with a source training dataset 
${\cal D}$$=$$\left\{ \left(\mathbf{x}_{i},\mathbf{y}_{i}\right)\right\} $, 
where $\mathbf{x}_{i}$ is an image and each label $y_i$ consists of object categories and bounding box coordinates. We consider scenarios in which there exists a covariate shift between the
input distribution $p_{S}:{\cal X}\times{\cal Y}\rightarrow\mathbb{R}^{+}$
of the original source data ${\cal D}$ and the target test distribution
$p_{T}:{\cal \overline{X}}\times{\cal Y}\rightarrow\mathbb{R}^{+}$.
More formally, we assume that $p_{S}\left(\mathbf{y}\mid\mathbf{x}\right)=p_{T}\left(\mathbf{y}\mid\mathbf{x}\right)$
but $p_{S}\left(\mathbf{x}\right)\neq p_{T}\left(\mathbf{x}\right)$
\cite{covariate-shift}.

\vspace{-1pt}
In the unsupervised domain adaptation setting, we are also given a set of unlabeled images ${\cal \overline{D}}=\left\{ \left(\overline{\mathbf{x}}_{j}\right)\right\} $
from the target domain, which we can use during training. Therefore, our objective
is to update the model parameters $\theta^{s}_{\tt M}$ into $\theta^{a}_{\tt M}$ to achieve a good performance on both the source test set and a given target test set, i.e., improving its robustness to the domain shifts. To effectively exploit the additional information in ${\cal \overline{D}}$,
we need to tackle two inter-related issues. First, the target training
set ${\cal \overline{D}}$ does not come with ground-truth labels. Second,
generating pseudo-labels for ${\cal \overline{D}}$ with the source model $\theta^{s}_{\tt M}$ 
leads to noisy supervision due to the domain shift and hinders the adaptation. In the following subsections, we present a simple approach for tackling these technical issues.

\vspace{-2pt}
\subsection{Simple adaptation for Robust Object Detection}\label{prop-framework}
\vspace{-6pt}
We present our simple adaptation method SimROD for enabling robust object detection models. SimROD integrates a teacher-guided fine-tuning, a new DomainMix augmentation method and a gradual adaptation technique. Sec. \ref{teacher-guided} describes the overall method. Next, Sec. \ref{domain-mix} presents the DomainMix augmentation, which is used for adapting both the teacher and student. Finally, Sec. \ref{gradual} explains the gradual adaptation that overcomes the two interrelated issues of domain shift and pseudo-label noise.

\vspace{-6pt}
\subsubsection{Overall approach}\label{teacher-guided}
\vspace{-5pt}

Our simple approach is motivated by the fact that label noise is exacerbated by the domain shift. Therefore, our approach aims to generate accurate pseudo-labels on target domain images and use them together with mixed images from source and target domain so as to provide strong supervision for adapting the models. 

Because the student target model may not have the capacity to generate accurate pseudo-labels and adapt itself, we propose to adapt an auxiliary teacher model first, which can later generate high-quality pseudo-labels for fine-tuning the student model. A flow diagram of SimROD is provided in Figure \ref{sys-diagram}. Its steps are summarized as follows: 

\vspace{-0pt}
\paragraph{Step 1:} We train a large source teacher model $\tt T$ with bigger capacity than the student model $\tt M$ to be adapted using the source data $\cal D$ and get parameters $\theta_{\tt T}^s$. The source teacher is used to generate initial pseudo-labels on target data.
\vspace{-10pt}
\paragraph{Step 2:} We adapt the large teacher model parameters from $\theta_{\tt T}^s$ to $\theta_{\tt T}^a$ using the gradual adaptation of Algorithm \ref{algo:proposed} (see Sec. \ref{gradual}). During this step, we use mixed images generated by the DomainMix augmentation (see Sec. \ref{domain-mix})
\vspace{-10pt}
\paragraph{Step 3:} We refine the pseudo-labels on the target data $\cal \overline{D}$ using the adapted teacher model parameters $\theta_{\tt T}^a$. Then, we fine-tune the student model $\tt M$ using these pseudo-labels in line 2 and 8 of Algorithm \ref{algo:proposed}.
\vspace{4pt}

One benefit of this approach is that it can adapt both small and large object detection models to domain shifts since it produces high quality pseudo-labels even when the student network is small. Another advantage of our method is that the teacher and student do not need to share the same architecture. Thus, it is possible to use a slow but accurate teacher for the purpose of adaptation while choosing a fast architecture for deployment. 

\vspace{-8pt}
\subsubsection{DomainMix augmentation}\label{domain-mix}
\vspace{-6pt}

\begin{figure}
\centering\includegraphics[width=0.25\paperwidth]{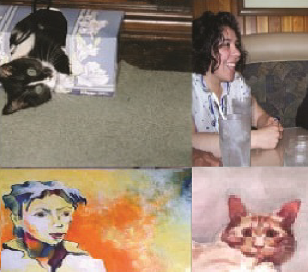}\vspace{-2pt}
\caption{\small An example image generated by DomainMix mixing real images from Pascal VOC and artistic images from Watercolor2K.}
\label{fig:domainmix_example}
\end{figure}

Here, we present a new augmentation method named DomainMix. As illustrated in Figure \ref{sys-diagram}, it uniformly samples images from both the source and target domains {${\cal D} \cup \overline{{\cal D}}$} and strongly mixes these images into a new image along with their (pseudo-)labels. Figure~\ref{fig:domainmix_example} shows an example of DomainMix images from natural and artistic domains.

DomainMix uses simple ideas with many benefits to mitigate domain shift and label noise:
\begin{itemize}[leftmargin=*]
    \item It produces a diverse set of images by randomly sampling and mixing crops from source and target sets with replacement. As a result, it uses a different sample of images at every epoch, thus increasing the effective number of training samples and preventing overfitting. In contrast, simple batching reuses same images at every epoch.
    \item It is data-efficient as it uses a weighted balanced sampling from both domains. This helps learning representations that are robust to data shifts even if the target dataset has limited samples or the source and target datasets are highly imbalanced. In \cite{Authors21supplementary}, we provide ablation studies that demonstrate the data efficiency of DomainMix.
    \item It mixes ground-truth and pseudo-labels in the same image. This mitigates the effect of false labels during adaptation because the image always contains accurate labels from the source domain
    \item It enforces the model to detect small objects as the objects in original samples are scaled down. 
\end{itemize}

\begin{algorithm}[ht]
\caption{\label{algo:domainmix} DomainMix augmentation} 
\begin{algorithmic}[1]
\algnewcommand\algorithmicinput{\textbf{Inputs:}}
\algnewcommand\INPUT{\item[\algorithmicinput]}
\algnewcommand\algorithmicoutput{\textbf{Output:}}
\algnewcommand\OUTPUT{\item[\algorithmicoutput]} 

\INPUT A batch $\beta$ of {$B$} images, labels {$\{\mathbf{y}_{i}\}$} from source data {$\cal D$}, unlabeled target data {$\cal \overline{D}$}, pseudo-labels {$\{\overline{\mathbf{y}}_{j}\}$}

\OUTPUT A batch of domain-mixed samples {$\widehat{\beta}$}
\Procedure{DomainMix}{$\beta,{\cal \overline{D}}, \{\overline{\mathbf{y}}_{j}\}$}

\State $\widehat{\beta} \gets \emptyset$

\For{$i\gets 1,B$}
    \State ${\cal S} \gets \{(\mathbf{x}_{i}, \mathbf{y}_{i})\}$
	\For{$j\gets \text{sample}({\cal D} \cup \overline{{\cal D}}, 3)$}
	    \If{$j \in \overline{\cal D}$}
	        \State {${\cal S} \gets {\cal S} \cup {\{(\overline{\mathbf{x}}_{j},\overline{\mathbf{y}}_{j})\}}$}
	    \Else
	        \State {${\cal S} \gets {\cal S} \cup \{(\mathbf{x}_{j},\mathbf{y}_{j})\}$}
	    \EndIf
	\EndFor
    \State Collate crops from 4 images in $\cal S$ into $\widehat{\mathbf{x}}_{i}$
    \State Recompute all box coordinates in $\cal S$ into $\widehat{\mathbf{y}}_{i}$
    \State $\widehat{\beta} \gets \widehat{\beta} \cup \{(\widehat{\mathbf{x}}_{i},\widehat{\mathbf{y}}_{i})\}$
\EndFor
\EndProcedure

\end{algorithmic}
\end{algorithm}

The steps of DomainMix augmentation are listed in Algorithm \ref{algo:domainmix}. For each image in a batch, we randomly sample three additional images from source and target data ${\cal D} \cup \overline{{\cal D}}$ and mix random crops of these images to create a new domain-mixed image in a $2\times2$ collage. In addition, we collate the pseudo-labels $\overline{\mathbf{y}}_{j}$ for the unlabeled examples ${\overline{\mathbf{x}}_{j}}$ in {$\cal \overline{D}$} with the ground-truth labels of source images. The bounding box coordinates of the objects are computed based on the relative position of each crop in the new mixed image. Furthermore, we employ a weighted balanced sampler to sample uniformly from the two domains.

\vspace{-8pt}
\subsubsection{Gradual self-labeling adaptation}\label{gradual}
\vspace{-6pt}

\begin{algorithm}
\caption{\label{algo:proposed}Gradual self-labeling adaptation}
\begin{algorithmic}[1]
\algnewcommand\algorithmicinput{\textbf{Inputs:}}
\algnewcommand\INPUT{\item[\algorithmicinput]}
\algnewcommand\algorithmicoutput{\textbf{Output:}}
\algnewcommand\OUTPUT{\item[\algorithmicoutput]} 

\INPUT Source model {$\theta^{s}_{\tt M} $}, labeled source data {$\cal D$}, unlabeled target data {$\cal \overline{D}$}, warmup epochs {$w$}, total epochs T, steps per epoch {$N$}, and batch size {$B$}
\OUTPUT Adapted model {$\theta^{a}_{\tt M} $}
\Procedure{Adapt}{$\theta^{s}_{\tt M},{\cal D},{\cal \overline{D}}$}

\For{$\overline{\mathbf{x}}_{j}\gets \cal \overline{D}$}
	{$\overline{\mathbf{y}}_{j}\gets \text{GenPseudo}(\overline{\mathbf{x}}_{j}, \theta^{s}_{\tt M})$}
\EndFor

\State Initialize $\theta\gets {\theta^{s}_{\tt M}}$

\For{$layer\gets \theta.\text{layers}$}
	\If{layer is not BatchNorm}
		Freeze layer
	\EndIf
\EndFor

\For{$epoch\gets 1,\ldots, T$}
	\If{epoch == w}  \Comment{switch to Phase 2}
		\For{$\overline{x}\gets \cal \overline{D}$}
			{$\overline{\mathbf{y}}_{j}\gets \text{GenPseudo}(\overline{\mathbf{x}}_{j}, \theta)$}
		\EndFor
		\State Unfreeze all layers
	\EndIf
	
	\For{$\text{step}\gets 1,\ldots, N$}
		\State Sample a batch $\beta=\{(\mathbf{x}_{i},\mathbf{y}_{i})\}_{i=1}^{B}$  from $\cal D$
		\State {$\widehat{\beta} \gets \text{DomainMix}(\beta,{\cal \overline{D}}, \{\overline{\mathbf{y}}_{j}\})$} as in Algo \ref{algo:domainmix}.
		\State Update {$\theta $} to minimize the loss with $\widehat{\beta}$
	\EndFor
\EndFor
\State ${\theta^{a}_{\tt M}\gets \theta}$
\EndProcedure

\end{algorithmic}
\end{algorithm}

Next, we present a gradual adaptation for optimizing the parameters of the detection model. This algorithm mitigates the effects of label noise, which is exacerbated by the domain shift. In fact, the pseudo-labels generated by the source models can be noisy on target domain images (e.g. it cannot detect objects or detects them inaccurately). If these initial pseudo-labels are used to adapt all the layers of the model at the same time, it results in poor supervision and hinders the model adaptation.

Instead, we propose a phased approach. First, we freeze all convolutional layers and adapts only the BN layers in the first $w$ epochs. After this first phase, BN layers' trainable coefficients are updated. The partially adapted model is then used to generate more accurate pseudo-labels, which is done offline for simplicity. In the second phase, all layers are unfrozen and then fine-tuned using the refined pseudo-labels. Note that during these two phases, we use the mixed image samples generated by the DomainMix augmentation. The detailed steps of this gradual adaptation are listed in Algorithm~\ref{algo:proposed}. 

In contrast to prior works \cite{adabn, bn-schneider20}, which used BN Adaption on its own, we integrate it within a self-training framework to effectively overcome the inevitable label noise caused by the domain shift \cite{cross-domain}. As will be shown in Section~\ref{experiments}, when used with the DomainMix augmentation, the resulting method is effective in adapting object detection models to different kinds of domain shifts.

Note that \cite{cross-domain} also used a two-phase progressive adaptation method but they used synthetic domain-mapped images, which are generated by a conditional GAN, to fine-tune the model in the first phase. In contrast, our method leverages actual target domain images, which are mixed with source domain images using DomainMix augmentation, during the entire adaptation process.

\section{Experiments results}\label{experiments}
\vspace{-4pt}In this section, we evaluate the effectiveness of SimROD to combat different kinds of domain shifts, compare the performance with prior works on standard benchmarks, and conduct ablation studies. For our experiments, we adopted the single-stage detection  architecture Yolov5 \cite{yolov5} and used different model sizes by scaling the input size, width and depth.
We study synthetic-to-real and camera-setup shifts~\cite{DAF} in Section~\ref{city-shifts}, cross-domain artistic shifts~\cite{cross-domain} in Section~\ref{cross-result}, and robustness against image corruptions~\cite{Stylize} in Section~\ref{corrupt-result}. Training details and additional results are provided in the supplementary materials~\cite{Authors21supplementary}.


\subsection{Synthetic-to-real and cross-camera benchmark}\label{city-shifts}
\vspace{-6pt}
\textbf{Datasets.} We used Sim10k \cite{SIM10K} to Cityscapes \cite{Cityscapes} and KITTI \cite{KITTI} to Cityscapes benchmarks to study the ability to adapt in synthetic-to-real and cross-camera shifts, respectively. Following prior works, only the \emph{car} class was used.

\begin{table*}
\centering
\fontsize{8}{10}\selectfont
\begin{tabular}[t]{lclcccrcl}
\toprule
Method & Arch. & Backbone & Source & AP50 & Oracle & $\tau$ & $\rho$ & Reference\\
\midrule
\addlinespace[0.3em]
DAF~\cite{DAF} & F-RCNN & V & 30.10 & 39.00 & - & 8.90 & - & CVPR 2018\\
MAF~\cite{MAF} & F-RCNN & V & 30.10 & 41.10 & - & 11.00 & - & ICCV 2019\\
RLDA~\cite{RLDA} & F-RCNN & I &  \bftab 31.08 &  \bftab 42.56 & 68.10 &  \bftab 11.48 &  \bftab 31.01 & ICCV 2019\\
\midrule
SCDA~\cite{SCDA} & F-RCNN & V & 34.00 & 43.00 & - & 9.00 & - & CVPR 2019\\
MDA~\cite{MDA} & F-RCNN & V & 34.30 & 42.80 & - & 8.50 & - & ICCV 2019\\
SWDA~\cite{SWDA} & F-RCNN & V & 34.60 & 42.30 & - & 7.70 & - & CVPR 2019\\
Coarse-to-Fine~\cite{C2F} & F-RCNN & V &  \bftab 35.00 & 43.80 & 59.90 & 8.80 & 35.34 & CVPR 2020\\
SimROD (self-adapt) & YOLOv5 & S320 & 33.62 & 38.73 & 48.81 & 5.11 & 33.66 & Ours\\
SimROD (w. teacher X640) & YOLOv5 & S320 & 33.62 &  \bftab 44.70 & 48.81 &  \bftab  11.08 &  \bftab 72.93 & Ours\\
\midrule
MTOR~\cite{MTOR} & F-RCNN & R & 39.40 & 46.60 & - & 7.20 & - & CVPR 2019\\
EveryPixelMatters~\cite{Every} & FCOS & V &  \bftab 39.80 & 49.00 & 69.70 & 9.20 & 30.77 & ECCV 2020\\
SimROD (self adapt) & YOLOv5 & S416 & 39.57 & 44.21 & 56.49 & 4.63 & 27.37 & Ours\\
SimROD (w. teacher X1280) & YOLOv5 & S416 & 39.57 &  \bftab 52.05 & 56.49 &  \bftab 12.47 &  \bftab 73.73 & Ours\\
\bottomrule
\end{tabular}
\caption{Results of different method/model pairs for the Sim10K-to-Cityscapes adaptation scenario. “V”, “I” and “R” represent the VGG16, ResNet50, Inception-v2 backbones respectively. "S320", “M416”,  “X640”, “X1280” represent different scales of Yolov5 model with increasing depth, width and input size. “Source” refers to the model trained only using source images without domain adaptation. For a fair comparison, we group together method/model pairs whose “Source” performance are similar. We report the AP50 (\%) performance of the adapted model and the “Oracle” model which is trained with labeled target data, as well each method's absolute and effective gains (\%) when available. $\tau$ and $\rho$ are the absolute gain and the effective gain respectively as defined in (\ref{tau}) and (\ref{eq:rho}).}
\label{sim10k_table}
\end{table*}

\textbf{Metrics.} For a fair comparison, we grouped different model/method pairs whose “Source” models (trained only on the labeled source data) have a similar average precision $\text{AP}^{50}(\theta^s)$ on the target test set (i.e. Cityscapes val). 
We compared each group based on three metrics: (1) $\text{AP}^{50}(\theta^a)$ of their “Adapted” models, (2) absolute adaptation gains $\tau$, and (3) their effective adaptation gains $\rho$ defined as:
\begin{align}
\tau & = \text{AP}^{50}(\theta^a)-\text{AP}^{50}(\theta^s) \label{tau}, \\ 
\rho & = 100\times \frac{\text{AP}^{50}(\theta^a)-\text{AP}^{50}(\theta^s)}{\text{AP}^{50}(\text{Oracle})-\text{AP}^{50}(\theta^s)}, \label{eq:rho}
\end{align}
where “Oracle” is the model that is trained with the labeled target domain data. The gain metric $\tau$ was proposed by \cite{C2F} to compare methods that may share same base architecture but have different performance before adaptation. For a better comparison, we also analyze the effectiveness of the adaptation method using the metric $\rho$. This metric helps understand if an adaptation method offers higher performance on the target test set beyond what is expected from having high performance on the source test set. A method that fails to adapt a model will have an effective gain of $\rho=0\%$ for that model whereas a method that gives a target performance close to the Oracle will have $\rho=100\%$.

\begin{figure}
\centering\includegraphics[width=1.\columnwidth]{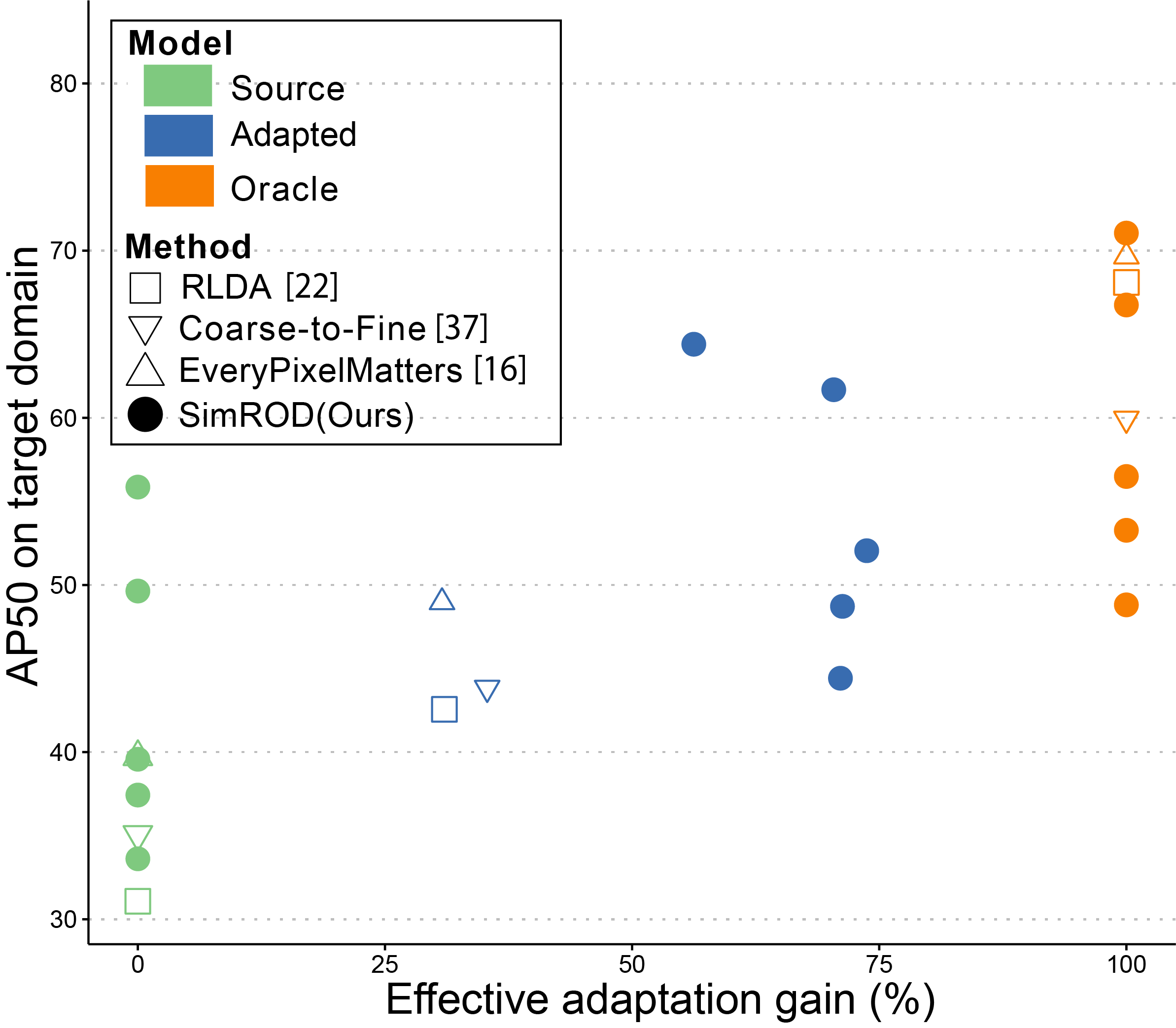}\vspace{-4pt}
\caption{\small AP50 on test vs effective gain for Sim10K to Cityscapes. We use five different backbones S320, M320, S416, S640 and M640 for the student and the same backbone X1280 for teacher.}
\label{fig:sim10k_gain_to_ap}
\end{figure}

\textbf{Sim10K to Cityscapes.} Table \ref{sim10k_table} shows that SimROD achieved new SOTA results on both the target AP50 performance and on the effective adaptation gain. We use two student models S320 and S416, which have the same Yolov5s architecture but different input sizes of 320 and 416 pixels to compare with prior methods that have comparable Source AP50 performance. For example, our S320 models achieves $AP50=44.70\%$ and $\rho=72.93\%$ compared to $AP50=43.8\%$ and $\rho=35.34\%$ for Coarse-to-Fine \cite{C2F}. Similar results were observed when comparing the performance of our adapted S416 model with that of the FCOS model adapted with EPM \cite{Every}.
Fig. \ref{fig:sim10k_gain_to_ap} demonstrates the effectiveness of SimROD to adapt models from Sim10K to Cityscapes compared to prior baselines. Models adapted with SimROD enjoyed up to 70-75\% of the target AP performance (that is obtained if the model was trained with a fully labeled target dataset). In contrast, the baseline methods achieved only about 30\% of their Oracle performance.

\textbf{KITTI to Cityscapes benchmark.} Table \ref{kitti_table} shows the results of this experiment, where SimROD outperformed the baselines. With the S416 model, it achieves slightly higher AP50 performance than the best baseline PDA \cite{PDA}. When using the medium size M416 model, SimROD also outperformed prior baselines with comparable Source AP50 performance namely SCDA \cite{SCDA} and EPM \cite{Every}.

\begin{table*}
\fontsize{8}{10}\selectfont
\centering
\begin{tabular}[t]{lllcccrcl}
\toprule
Method & Arch. & Backbone & Source & AP50 & Oracle & $\tau$ & $\rho$ & Reference\\
\midrule
DAF~\cite{DAF} & F-RCNN & V & 30.20 & 38.50 & - & 8.30 & - & CVPR 2018\\
MAF~\cite{MAF} & F-RCNN & V & 30.20 & 41.00 & - & 10.80 & - & ICCV 2019\\
RLDA~\cite{RLDA} & F-RCNN & I & 31.10 & 42.98 & 68.10 & 11.88 & 32.11 & ICCV 2019\\
PDA~\cite{PDA} & F-RCNN & V & 30.20 & 43.90 & 55.80 & 13.70 & 53.52 & WACV 2020\\
SimROD (self-adapt) & YOLOv5 & S416 & 31.61 & 35.94 & 56.15 & 4.33 & 17.65 & Ours\\
SimROD (w. teacher X1280) & YOLOv5 & S416 & \bftab 31.61 & \bftab 45.66 & 56.15 & \bftab 14.05 & \bftab 57.27 & Ours\\
\midrule
SCDA~\cite{SCDA} & F-RCNN & V & \bftab 37.40 & 42.60 & - & 5.20 & - & CVPR 2019\\
EveryPixelMatters~\cite{Every} & FCOS & R & 35.30 & 45.00 & 70.40 & 9.70 & 27.64 & ECCV 2020\\
SimROD (self adapt) & YOLOv5 & M416 & 36.09 & 42.94 & 59.29 & 6.85 & 29.51 & Ours\\
SimROD (w. teacher X1280) & YOLOv5 & M416 & 36.09 & \bftab 47.52 & 59.29 & \bftab 11.43 & \bftab 49.26 & Ours\\
\bottomrule
\end{tabular}
\caption{Results of different method/model pairs on the KITTI-to-Cityscapes adaptation scenario. $\tau$ and $\rho$ are the absolute gain and the effective gain respectively as defined in (\ref{tau}) and (\ref{eq:rho}).}
\label{kitti_table}
\end{table*}


\vspace{-0pt}
\subsection{Cross-domain artistic benchmark}\label{cross-result}
\vspace{-4pt}

\begin{table*}
\centering
\fontsize{8}{10}\selectfont
\begin{tabular}[t]{lllrrrrrl}
\toprule
Method & Arch. & Backbone & Source & AP50 & Oracle & $\tau$ & $\rho$ & Reference\\
\midrule
DAF~\cite{DAF} & F-RCNN & V & 39.80 & 34.30 & NA & -5.50 & NA & CVPR 2018\\
DAM~\cite{DAM} & F-RCNN & V & \bftab 39.80 & 52.00 & NA & 12.20 & NA & CVPR 2019\\
DeepAugment~\cite{DeepAugment} & YOLOv5 & S416 & 37.46 & 45.19 & 56.07 & 7.73 & 41.54 & arXiv 2020\\
{BN-Adapt}~\cite{bn} & YOLOv5 & S416 & 37.46 & 45.72 & 56.07 & 8.26 & 44.39 & NeurIPS 2020\\
Stylize~\cite{Geirhos19} & YOLOv5 & S416 & 37.46 & 46.26 & 56.07 & 8.80 & 47.29 & arXiv 2019\\
STAC~\cite{STAC} & YOLOv5 & S416 & 37.46 & 49.83 & 56.07 & 12.37 & 66.47 & arXiv 2020\\
DT+PL~\cite{cross-domain} & YOLOv5 & S416 & 37.46 & 44.86 & 56.07 & 7.40 & 39.77 & CVPR 2018\\
SimROD (self-adapt) & YOLOv5 & S416 & 37.46 & 52.58 & 56.07 & 15.12 & 81.26 & Ours\\
SimROD (teacher X416) & YOLOv5 & S416 & 37.46 & \bftab 55.55 & 56.07 & \bftab 18.09 & \bftab 97.21 & Ours\\
\midrule
ADDA~\cite{ADDA} & SSD & V & 49.60 & 49.80 & 58.40 & 0.20 & 2.27 & CVPR 2017\\
DT+PL~\cite{cross-domain} & SSD & V & \bftab 49.60 & 54.30 & 58.40 & 4.70 & 53.41 & CVPR 2018\\
SWDA~\cite{SWDA} & F-RCNN & V & 44.60 & 56.70 & 58.60 & 12.10 & \bftab 86.43 & CVPR 2019\\
DeepAugment~\cite{DeepAugment} & YOLOv5 & M416 & 46.95 & 54.02 & 66.34 & 7.07 & 36.47 & arXiv 2020\\
{BN-Adapt}~\cite{bn} & YOLOv5 & M416 & 46.95 & 55.75 & 66.34 & 8.80 & 45.39 & NeurIPS 2020\\
Stylize~\cite{Geirhos19} & YOLOv5 & M416 & 46.95 & 55.24 & 66.34 & 8.29 & 42.76 & arXiv 2019\\
STAC~\cite{STAC} & YOLOv5 & M416 & 46.95 & 57.82 & 66.34 & 10.87 & 56.07 & arXiv 2020\\
DT+PL~\cite{cross-domain} & YOLOv5 & M416 & 46.95 & 49.14 & 66.34 & 2.19 & 11.30 & CVPR 2018\\
SimROD (self-adapt) & YOLOv5 & M416 & 46.95 & 60.08 & 66.34 & 13.13 & 67.72 & Ours\\
SimROD (teacher X416) & YOLOv5 & M416 & 46.95 & \bftab 63.47 & 66.34 & \bftab 16.52 &  85.22 & Ours\\
\bottomrule
\end{tabular}
\caption{\small Benchmark results on Real (VOC) to Watercolor2K domain shift.}
\label{tab:water}
\end{table*}

\textbf{Datasets and metrics.} The cross-domain artistic benchmark consists of three domain shifts where the source data is VOC07 trainval and the target domains are Clipart1k, Watercolor2k and Comic2k datasets \cite{cross-domain}. We use the same benchmark metrics as in Sec. \ref{city-shifts}.

\textbf{Results.} Our method outperformed the baselines by significant margins. Compared to DT+PL \cite{cross-domain}, our method further improved the AP50 of the yolov5s model by absolute gains of +8.45, +12 and +10.69 \% points on Clipart, Comic, and Watercolor respectively. While DT+PL outperformed the augmentation-based baselines on Clipart, it did slightly worse than STAC on Comic and Watercolor. Finally, SimROD was effective in adapting models of different sizes. Without generating synthetic data or using domain adversarial training, SimROD's effective gain $\rho$ was consistently above 70\% and could reach up to 97\% when a large adapted teacher was used to refine the pseudo-labels.

In Table \ref{tab:water}, we give a detailed benchmark for the VOC to Watercolor benchmark, from which we used 1000 unlabeled images as target data. In \cite{Authors21supplementary}, we present detailed results on Clipart and Comic dataset as well as more ablation results when using extra unlabeled data for adaptation.

\begin{figure*}
\centering\includegraphics[width=2.0\columnwidth]{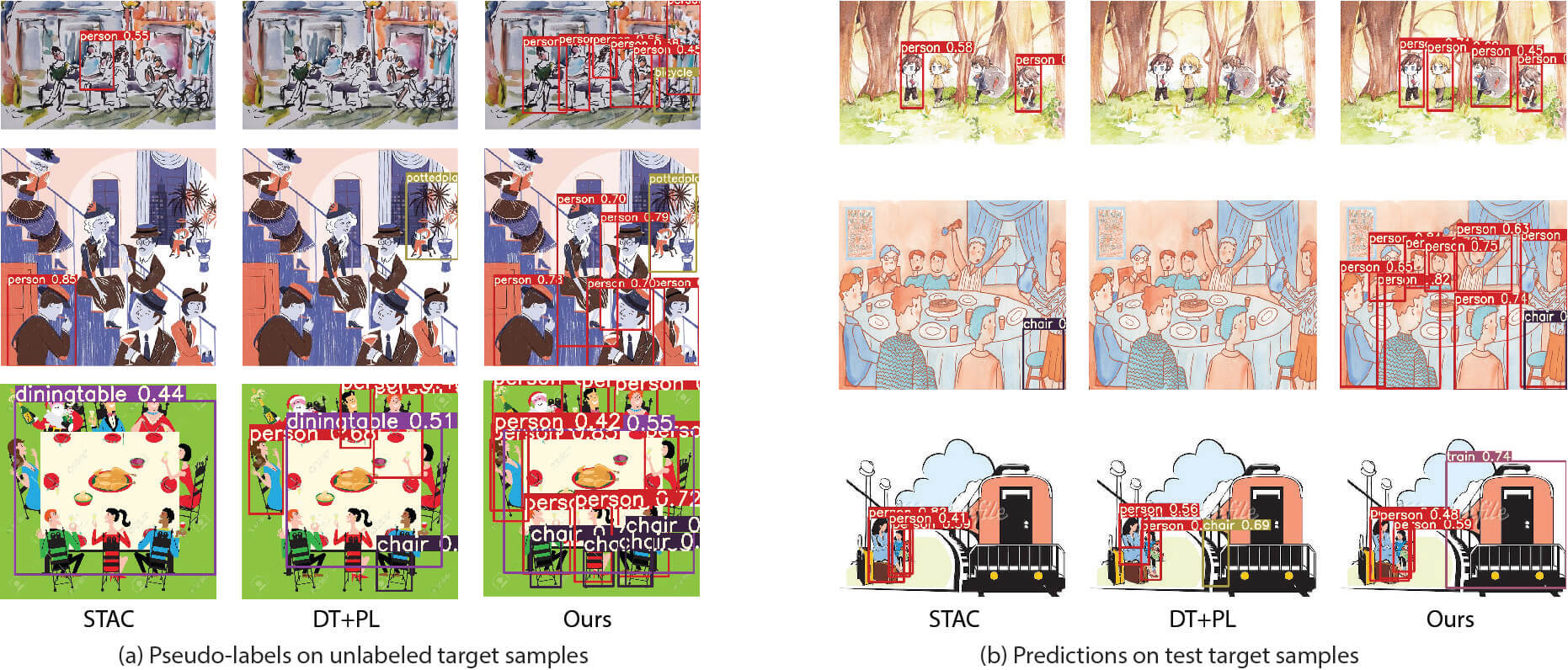}\vspace{-4pt}
\caption{\small Qualitative comparison: (a) pseudo-labels generated on unlabeled target examples and (b) test predictions with adapted Yolov5s.}
\label{qualitative}
\end{figure*}


\vspace{0pt}
\subsection{Image corruptions benchmark}\label{corrupt-result} 
\vspace{-4pt}

\textbf{Datasets.} We evaluate our method's robustness to image corruption using the standard benchmarks Pascal-C, COCO-C, and Cityscapes-C \cite{Stylize}. For Pascal-C, we used VOC07 trainval split as the source training data. For COCO-C and Cityscapes-C, we divided the train split and used the first half as source training data. There are $N_{c}=15$ different corruption types for each dataset. Thus, we applied each corruption type on the VOC12 trainval or on the second half of COCO-C and Cityscapes-C train as unlabeled target data. Precisely, we applied each corruption type with middle severity onto each image using the \emph{imagecorruptions} library \cite{Stylize}. More details are given in \cite{Authors21supplementary}.

\textbf{Metrics.} For image corruption benchmark, we followed the evaluation protocol from \cite{hendrycks2019benchmarking, Stylize, taori2020} and measured the mean performance under corruption (mPC), relative performance under corruption (rPC), and the relative robustness $\tau_c$ of the adapted model averaged over $N_{c}$ corruption types:  
\vspace{-12pt}
\begin{gather}
\text{mPC}^{x} =\frac{1}{N_{c}}\sum_{c=1}^{N_{c}}\frac{1}{N_{s}}\sum_{s=1}^{5}\text{AP}_{c,s}^{x}.\\
\text{rPC}^{x} =\frac{\text{mPC}^{x}}{\text{AP}_{\text{clean}}^{x}}.\\
\tau_c = \text{mPC}(\theta^a)-\text{mPC}(\theta^s).
\vspace{-4pt}
\end{gather}
where $\text{AP}_{\text{clean}}^{x}$ and $\text{AP}_{c,s}^{x}$ denote the average precision of the test data with corruption type $c$ and severity level $s$. The relative robustness $\tau_c$ quantifies the effect of adaptation on the performance under distribution shift (mPC).

\textbf{Baselines.} We use the following baselines which were proposed to improve the robustness to image corruptions: Stylize~\cite{Geirhos19}, BN-Adapt~\cite{bn}, DeepAugment~\cite{DeepAugment}, STAC~\cite{STAC}, and DT+PL~\cite{cross-domain}. Unless specified, we employed weak data augmentations such as RandomHorizontalFlip and RandomCrop for all baselines.

\textbf{Main results.} Table \ref{tab:result-pascal-c}, \ref{tab:result-coco-c} and \ref{tab:result-city-c} show the results of Yolov5m model for Pascal-C, COCO-C, and Cityscapes-C, respectively. We report the results with different model sizes in \cite{Authors21supplementary}. We used the large model Yolov5x model as a teacher. An ablation study on Pascal-C is provided in Table \ref{tab:ablation} and will be discussed later. 

\begin{table}
\centering
\fontsize{8}{10}\selectfont
\begin{tabular}[t]{lrrrrr}
\toprule
Method & $\text{AP}^{50}_{\text{clean}}$ & $\text{mPC}^{50}$ & $\text{rPC}$ & $\tau_c$ & $\rho$\\
\midrule
Source & 83.13 & 53.78 & 64.69 & 0.00 & 0\\
Stylize & 84.79 & 62.92 & 74.21 & 9.14 & 36.62\\
BN-Adapt & 83.01 & 64.60 & 77.82 & 10.82 & 43.35\\
DeepAugment & 85.05 & 64.88 & 76.28 & 11.10 & 44.47\\
STAC & \bftab 87.00 & 66.88 & 76.87 & 13.10 & 52.48\\
\bftab SimROD (ours) & 86.97 & \bftab 75.40 & \bftab 86.70 & \bftab 21.62 & \bftab 86.62\\
\midrule
Oracle & 86.75 & 78.74 & 90.77 & 24.96 & 100 \\
\bottomrule
\end{tabular}
\caption{\label{tab:result-pascal-c} \small Performance comparison on Pascal-C benchmark.}
\end{table}

\textbf{Unlabeled target samples improved robustness to image corruption.} The source models suffered from performance drop due to image corruptions. By adapting the models with SimROD, the mean performance under corruption $\text{mPC}^{50}$ was significantly improved by +21.62, +6.43, and +6.48 absolute percentage points on Pascal-C, COCO-C, and Cityscapes-C, respectively. Our method outperformed the Stylize, DeepAugment, BNAdapt baselines on all metrics. In fact, STAC, which also used unlabeled target samples, achieved the second best performance. This shows that augmentation or batch norm adaptation is not sufficient to fix the domain shift on all possible corruptions. Instead, using unlabeled samples from target domain is more effective to combat image corruptions.

\textbf{Pseudo-label refinement ensured performance close to Oracle}. Moreover, Tables \ref{tab:result-pascal-c}, \ref{tab:result-coco-c} and \ref{tab:result-city-c}  show that the performance of our unsupervised method was close to that of the Oracle, which uses ground-truth labels for \textit{target} domain data. This was possible because the adapted teacher produces highly accurate pseudo-labels, which could be used along with DomainMix augmentation to effectively adapt the student model.

\begin{table}
\centering
\fontsize{8}{10}\selectfont
\begin{tabular}[t]{lrrrrr}
\toprule
Method & $\text{AP}^{50}_{\text{clean}}$ & $\text{mPC}^{50}$ & $\text{rPC}$ & $\tau_c$ & $\rho$\\
\midrule
Source & \bftab 36.85 & 22.03 & 59.78 & 0.00 & 0\\
Stylize & 35.75 & 23.82 & 66.63 & 1.79 & 22.02\\
BN-Adapt & 36.24 & 24.79 & 68.41 & 2.76 & 33.95\\
DeepAugment & 35.51 & 24.33 & 68.52 & 2.30 & 28.29\\
STAC & 36.76 & 24.80 & 67.46 & 2.77 & 34.07\\
\bftab SimROD (ours) & 36.79 & \bftab 28.46 & \bftab 77.36 & \bftab 6.43 & \bftab 79.09\\
\midrule
Oracle & 36.23 & 30.16 & 83.25 & 8.13 & 100\\
\bottomrule
\end{tabular}
\caption{\label{tab:result-coco-c}\small Performance benchmark on COCO-C dataset.}
\end{table}

\begin{table}
\centering
\fontsize{8}{10}\selectfont
\begin{tabular}[t]{lrrrrr}
\toprule
Method & $\text{AP}^{50}_{\text{clean}}$ & $\text{mPC}^{50}$ & $\text{rPC}$ & $\tau_c$ & $\rho$\\
\midrule
Source & 19.48 & 11.53 & 59.19 & 0.00 & 0\\
Stylize & 21.77 & 14.62 & 67.16 & 3.09 & 25.81\\
DeepAugment & 20.28 & 14.79 & 72.93 & 3.26 & 27.23\\
STAC & \bftab 24.54 & 15.39 & 62.71 & 3.86 & 32.25\\
\bftab SimROD (ours) & 24.06 & \bftab 18.01 & \bftab 74.85 & \bftab 6.48 & \bftab 54.14\\
\midrule
Oracle & 26.58 & 23.50 & 88.41 & 11.97 & 100\\
\bottomrule
\end{tabular}
\caption{\label{tab:result-city-c}\small Performance benchmark on Cityscapes-C dataset.}
\end{table}

\begin{table}
\small\centering
\fontsize{8}{10}\selectfont
\begin{tabular}[t]{lccccll}
\toprule
Method & TG & DMX & GA & FT & $\text{mPC}^{50}$ & $\tau_c$\\
\midrule
\addlinespace[0.3em]
Source &  &  &  &  & 53.78 & 0.0\\
BN-Adapt &  &  & \checkmark &  & 64.60 & 10.8\\
BN-A + DMX &  & \checkmark & \checkmark &  & 66.78 & 13.0\\
SimROD w/o TG &  & \checkmark & \checkmark & \checkmark & 71.81 & 18.0\\
SimROD w/o GA & \checkmark & \checkmark &  & \checkmark & 73.45 & 19.7\\
SimROD & \checkmark & \checkmark & \checkmark & \checkmark & 75.40 & 21.7\\
\bottomrule
\end{tabular}
\vspace{-0pt}
\caption{\small Ablation study on Pascal-C with yolov5m. See~\cite{Authors21supplementary} for ablations with other models. TG, GA, DMX, and FT denote Teacher Guidance, Gradual Adaption, DomainMix, and Fine-Tuning.}
\label{tab:ablation}
\end{table}

\textbf{Ablation Study}. Next, we present an ablation study using the Yolov5m model on Pascal-C in Table \ref{tab:ablation} to gain some insights about the contributions of the three parts of our method. First, BN-Adapt improved the mean performance under corruption by 10.82\% AP50. Applying DomainMix augmentation on top of BN-Adapt improved the performance by 2.18\%. Next, the teacher-guided (TG) pseudo-label refinement was particularly useful in adapting small models. When using our full method, the performance increased by 10.8\% compared to BN-Adapt. Compared to self adaptation, TG improved the Yolov5 model's performance $\text{mPC}$ by +3.7 \%. Finally, the gradual adaptation (GA) also played an important role in refining pseudo-labels and in improving the model's robustness. For example, if we did not use GA and skipped the BN adaptation in the first phase, the performance dropped by 1.95\% compared to the full method. Our method organically integrates these parts to tackle UDA for object detection. While the parts may appear simple, their synergy helped mitigate the challenging issues of domain shift and pseudo-label noise.

\textbf{Qualitative analysis} Finally, we illustrate the effectiveness of our method by showing the pseudo-labels generated with our method on the unlabeled target training images on Comic dataset.  As seen in Figure \ref{qualitative}(a), our method generated highly accurate pseudo-labels despite the domain shift. In contrast, STAC and DT+PL generated sparse labels since they missed to detect many objects. The performance difference transferred to the quality of predictions on the test set as shown in Figure \ref{qualitative}(b).

\section{Conclusion}\label{conclusion}
\vspace{-6pt}
We proposed a simple and effective unsupervised method for adapting detection models under domain shift. Our self-labeling framework gradually adapted the model using a new domain-centric augmentation method and a teacher-guided finetuning. Our method achieved significant gains in terms of model robustness compared to existing baselines both for small and large models. Not only our method did mitigate the effect of domain shifts due to low-level image corruptions but also it could  adapt the models when presented with high-level stylistic differences between the source and target domains. Through ablation study, we got some insights on why gradual adaptation works and how the teacher-guided pseudo-label refinement can help adapt the models. We hope this simple method will guide future progress of robust object detection research.

\balance

{\small
\bibliographystyle{ieee_fullname}
\bibliography{references}
}

\clearpage
\section*{Supplementary materials}
\vspace{-4pt}
The following supplementary materials provide further details on training, on the results of the different benchmarks, and more qualitative analysis and visualizations.






\section{Experiments setup}

\subsection{Training details and hyperparameters}

We trained each model using a standard stochastic gradient descent (SGD) optimizer with momentum parameter 0.937 and weight decay $5e^{-4}$. We used warm-up and cosine decay rule for training.
For the NMS parameters, we used an IoU threshold of $0.65$ and an object confidence threshold of $0.001$. When generating pseudolabels, we used a higher confidence threshold of 0.4. We used the model definitions, as defined by the initial release of YOLOv5 \cite{yolov5} with last commit id `364fcfd7d'.  Finally, we used a generalized IoU loss (GIoU) for localization and a focal loss for the classification loss and objectness loss for training the models. 

To manage our experiments and make our results reproducible, we used the open-source tool Guildai \cite{guildai}. Most hyper-parameters (Momentum, NMS, etc.) were set as defaults in YOLOv5 repo \cite{yolov5}. We tuned only the learning rate for each dataset. The value of hyperparameters are configured in the `guild.yml' file. For the gradual adaptation procedure, we use a large enough number of epochs for Phase 1 to ensure the convergence of BN adaptation. We use a separate validation set to maintain the best checkpoint using the validation AP. Therefore, we initialize the phase 2 training with the best checkpoint of phase 1. It is also worth noting that our framework does not add new hyperparameters. 

When training the COCO source models or the Stylize and DeepAugment baselines, we followed the training procedure in YOLOv5 \cite{yolov5} and trained the model from scratch using 300 epochs and a learning rate of 0.01.  For Pascal and Cityscapes the source models were obtained through transfer learning from COCO pretrained weights using $100$ and $200$ epochs respectively. For that, we used learning rate of $4e^{-5}$ and batch size of $128$. When applying our adaptation method, we also fine-tuned the source model using same learning rate of $4e^{-5}$, a batch size of $128$ and 100 epochs for all models and target domains. 

We did not use multi-scale training to simplify our analysis. The same image input size was used during training, pseudo-label generation and evaluation. For Sim10K/KITTI to Cityscapes, we specify the input size used to train each student and teacher model in our results. For the artistic benchmark, we use the same input size of 416 for both student and teacher models. For the image corruption benchmark, we used the same input size of 416 for Pascal-C and COCO-C whereas we used a larger size of 640 for Cityscapes-C.

For the Stylize baseline, we applied only one style for each image to keep the dataset size the same, to ensure a fair comparison. We preserved the original image dimensions and disabled the cropping. Alpha was fixed to 1 to apply the highest strength of stylization.

   

\subsection{More details on datasets}

Table \ref{tab:Pascal-C-summary}, \ref{tab:COCO-C-summary}, \ref{tab:Cityscapes-C-summary}, and \ref{tab:cross-summary} show a summary of the data splits that we used as the source or clean split versus target or stylized/augmented split for each dataset. 
To make a fair comparison, we keep the total number of images in the entire training data to be the same for all methods.

For Pascal-C, COCO-C, and Cityscapes-C, we generated the corrupted test set by applying each corruption to the clean test with all five severity levels. For the cross-domain adaptation benchmark, we used the test split for Clipart, Watercolor, or Comic for measuring test AP on the target domain.
\begin{itemize}
    \item \textbf{Sim10K}: we use the SIM10k dataset as the labeled source training data and the training set of Cityscapes as unlabeled target data. The validation set of Cityscapes was used as target test set.
    \item \textbf{KITTI}: we use the training set of KITTI as the labeled source data and the training set of Cityscapes as unlabeled target data. The validation set of Cityscapes was used as target test set.
    \item \textbf{Clipart/Watercolor/Comic}: the datasets used in Source, DeepAugment, Stylize baselines for Clipart/Watercolor/Comic are exactly the same as those used for Pascal-C. Other than this, the train set of Clipart/Watercolor/Comic were used as the target domain dataset. 
    In DT+PL experiments, we first apply the domain transfer on the union of VOC2007 trainval and VOC2012 trainval. Then, we apply the DT step on the source model using the domain-transferred dataset. Finally, we apply the PL step on the output model of DT step using the train split of of Clipart/Watercolor/Comic. Note that we do not use the ground-truth labels but use the pseudo-labels instead.
    \item \textbf{Pascal-C}: we used VOC2007 trainval as the source and VOC2012 trainval as the target. For the DeepAugment baseline, we augmented VOC2012 train with the CAE method and VOC2012 val with the EDSR method. We used VOC2007 test as the clean test set.
    \item \textbf{COCO-C}: we split COCO train2017 into two approximately equal halves and used the first half as source, the second half as target. For DeepAugment, we divided COCO train2017 in three random splits and used them for the clean split, CAE split and EDSR split respectively. 
    COCO val2017 was used as clean test.
    \item \textbf{Cityscapes-C}: we split the source domain and target domain by city names. We carefully chose the cities for each domain so that source and target are of approximately equal size. Of all 18 cities in cityscapes-train, 9 cities: `cologne', `krefeld', `bremen', `darmstadt', `hanover', `aachen', `stuttgart', `jena', and `tubingen' were used as source data; the other 9 cities: `bochum', `ulm', `monchengladbach',  `weimar', `strasbourg', `zurich', `hamburg', `dusseldorf', and `erfurt' were used as target data.
    When training the DeepAugment baseline for Cityscapes, we further split the target domain into two splits. The first split that contains `zurich', `weimar', `erfurt', and `strasbourg' was augmented with the CAE method. The second split which contains `bochum', `ulm', `monchengladbach', `hamburg', `dusseldorf' was augmented with the EDSR method.
    The validation set of Cityscapes was used as clean test.

\end{itemize}

\begin{table*}
\centering
\fontsize{9}{11}\selectfont
\begin{tabular}{p{3.5cm}cc}
\toprule
Method & Source / Clean split (size) & Target / Augmented split (size) \\
\midrule
\hspace{0em} Source & VOC2007-trainval (5011) & N/A\\
\hspace{0em} DeepAugment & VOC2007-trainval (5011) & CAE VOC2012-train (5717) + EDSR VOC2012-val (5823) \\
\hspace{0em} Stylize & VOC2007-trainval (5011) & stylized VOC2012-trainval (11540) \\
\hspace{0em} BN-Adapt & VOC2007-trainval (5011) & VOC2012-trainval (11540) \\
\hspace{0em} STAC & VOC2007-trainval (5011) & VOC2012-trainval (11540) \\
\hspace{0em} SimROD (Ours) & VOC2007-trainval (5011) & VOC2012-trainval (11540) \\
\bottomrule
\end{tabular}
\vspace{-0pt}
\caption{\label{tab:Pascal-C-summary}Dataset splits used for Pascal-C}
\end{table*}

\begin{table*}
\centering
\fontsize{9}{11}\selectfont
\begin{tabular}{p{3.5cm}cc}
\toprule
Method & Source / Clean split (size) & Target / Augmented split (size) \\
\midrule
\hspace{0em} Source & coco-train2017/first half (58458) & N/A\\
\hspace{0em} DeepAugment & coco-train2017/first 1/3 (39088) & CAE second 1/3 (39088) + EDSR third 1/3 (39090)\\
\hspace{0em} Stylize & coco-train2017/first half (58458) & stylized coco-train2017/second half (58808) \\
\hspace{0em} BN-Adapt & coco-train2017/first half (58808) & coco-train2017/second half (58808) \\
\hspace{0em} STAC & coco-train2017/first half (58458) & coco-train2017/second half (58808) \\
\hspace{0em} SimROD (Ours) & coco-train2017/first half (58458) & coco-train2017/second half (58808) \\
\bottomrule
\end{tabular}
\vspace{-0pt}
\caption{\label{tab:COCO-C-summary}Dataset splits used for COCO-C}
\end{table*}

\begin{table*}
\centering
\fontsize{9}{11}\selectfont
\begin{tabular}{p{2.5cm}cc}
\toprule
Method & Source / Clean split (size) & Target / Augmented split (size) \\
\midrule
\hspace{0em} Source & cityscapes-train/first half (1483) & N/A\\
\hspace{0em} DeepAugment & cityscapes-train/first half (1483) & CAE train/second half-split 1 (732) + EDSR train/second half-split 2 (750) \\
\hspace{0em} Stylize & cityscapes-train/first half (1483) & stylized cityscapes-train/second half (1482) \\
\hspace{0em} Bn\_only & cityscapes-train/first half (1483) & cityscapes-train/second half (1482) \\
\hspace{0em} Stac & cityscapes-train/first half (1483) & cityscapes-train/second half (1482) \\
\hspace{0em} Ours w/o TG & cityscapes-train/first half (1483) & cityscapes-train/second half (1482)\\
\hspace{0em} Ours & cityscapes-train/first half (1483) & cityscapes-train/second half (1482) \\
\bottomrule
\end{tabular}
\vspace{-0pt}
\caption{\label{tab:Cityscapes-C-summary}Dataset splits used for Cityscapes-C}
\end{table*}

\begin{table*}
\centering
\fontsize{9}{11}\selectfont
\begin{tabular}{p{3.5cm}cc}
\toprule
Method & Source / Clean split (size) & Target / Augmented split (size) \\
\midrule
\hspace{0em} Source  & VOC2007-trainval (5011) & N/A\\
\hspace{0em} DeepAugment & VOC2007-trainval (5011) & CAE VOC2012-train (5717) + EDSR VOC2012-val (5823) \\
\hspace{0em} Stylize & VOC2007-trainval (5011) & stylized VOC2012-trainval (11540) \\
\hspace{0em} Bn\_only & VOC2007-trainval (5011) & clipart/watercolor/comic-train (500/1000/1000) \\
\hspace{0em} Stac & VOC2007-trainval (5011) & clipart/watercolor/comic-train (500/1000/1000) \\
\hspace{0em} Ours w/o TG & VOC2007-trainval (5011) & clipart/watercolor/comic-train (500/1000/1000)\\
\hspace{0em} Ours & VOC2007-trainval (5011) & clipart/watercolor/comic-train (500/1000/1000) \\
\bottomrule
\end{tabular}
\vspace{-0pt}
\caption{\label{tab:cross-summary}Dataset splits used for Clipart/Watercolor/Comic}
\end{table*}


\section{More results on synthetic-to-real and cross-camera benchmarks}

\subsection{Full results on Sim10K/KITTI to Cityscapes}

Table \ref{supp:sim10k_table} and \ref{supp:kitti_table} expand on the results reported in Table \ref{sim10k_table} and \ref{kitti_table} respectively. In particular, they show the performance of the teacher models and that of models adapted with the smaller teacher model X640.

\begin{table*}
\centering
\begin{tabular}{lclcccrcl}
\toprule
Method & Arch. & Backbone & Source & AP50 & Oracle & $\tau$ & $\rho$ & Reference\\
\midrule
\addlinespace[0.3em]
DAF~\cite{DAF} & F-RCNN & V & 30.10 & 39.00 & - & 8.90 & - & CVPR 2018\\
MAF~\cite{MAF} & F-RCNN & V & 30.10 & 41.10 & - & 11.00 & - & ICCV 2019\\
RLDA~\cite{RLDA} & F-RCNN & I &  \bftab 31.08 &  \bftab 42.56 & 68.10 &  \bftab 11.48 &  \bftab 31.01 & ICCV 2019\\
\midrule
SCDA~\cite{SCDA} & F-RCNN & V & 34.00 & 43.00 & - & 9.00 & - & CVPR 2019\\
MDA~\cite{MDA} & F-RCNN & V & 34.30 & 42.80 & - & 8.50 & - & ICCV 2019\\
SWDA~\cite{SWDA} & F-RCNN & V & 34.60 & 42.30 & - & 7.70 & - & CVPR 2019\\
Coarse-to-Fine~\cite{C2F} & F-RCNN & V &  \bftab 35.00 & 43.80 & 59.90 & 8.80 & 35.34 & CVPR 2020\\
SimROD (self-adapt) & YOLOv5 & S320 & 33.62 & 38.73 & 48.81 & 5.11 & 33.66 & Ours\\
SimROD (w. teacher X640) & YOLOv5 & S320 & 33.62 &  \bftab 44.70 & 48.81 &  \bftab  11.08 &  \bftab 72.93 & Ours\\
\midrule
MTOR~\cite{MTOR} & F-RCNN & R & 39.40 & 46.60 & - & 7.20 & - & CVPR 2019\\
EveryPixelMatters~\cite{Every} & FCOS & V &  \bftab 39.80 & 49.00 & 69.70 & 9.20 & 30.77 & ECCV 2020\\
SimROD (self adapt) & YOLOv5 & S416 & 39.57 & 44.21 & 56.49 & 4.63 & 27.37 & Ours\\
SimROD (w. teacher X640) & YOLOv5 & S416 & 39.57 & 51.68 & 56.49 & 12.10 & 71.53 & Ours\\
SimROD (w. teacher X1280) & YOLOv5 & S416 & 39.57 &  \bftab 52.05 & 56.49 &  \bftab 12.47 &  \bftab 73.73 & Ours\\
\midrule
SimROD (self-adapt) & YOLOv5 & M640 & 55.86 & 60.29 & 71.05 & 4.43 & 29.16 & Ours\\
SimROD (w. teacher X640) & YOLOv5 & M640 & 55.86 & 62.18 & 71.05 & 6.33 & 41.64 & Ours\\
SimROD (w. teacher X1280) & YOLOv5 & M640 & 55.86 & 64.40 & 71.05 & 8.54 & 56.24 & Ours\\
\midrule
SimROD (self-adapt) & YOLOv5 & X640 & 60.34 & 63.27 & 72.51 & 2.93 & 24.09 & Ours\\
SimROD (self-adapt) & YOLOv5 & X1280 &  71.66 &  75.94 & 82.90 & 4.28 & 38.08 & Ours\\
\bottomrule
\end{tabular}
\vspace{2pt}
\caption{Results of different method/model pairs for the Sim10K-to-Cityscapes adaptation scenario. “V”, “I” and “R” represent the VGG16, ResNet50, Inception-v2 backbones respectively. "S320", “M416”,  “X640”, “X1280” represent different scales of Yolov5 model with increasing depth, width and input size. “Source” denotes that the model is trained only using source images without domain adaptation. For fair comparison, we group together method/model pairs whose “Source” performance are similar. We report the AP50 (\%) performance of the adapted model and the “Oracle” model which is trained with labeled target data as well each method's absolute and effective gains (\%) when available. $\tau$ and $\rho$ are the absolute gain and the effective gain respectively as defined in (\ref{tau}) and (\ref{eq:rho}).}
\label{supp:sim10k_table}
\end{table*}

\begin{table*}
\centering
\begin{tabular}{lllcccrcl}
\toprule
Method & Arch. & Backbone & Source & AP50 & Oracle & $\tau$ & $\rho$ & Reference\\
\midrule
DAF~\cite{DAF} & F-RCNN & V & 30.20 & 38.50 & - & 8.30 & - & CVPR 2018\\
MAF~\cite{MAF} & F-RCNN & V & 30.20 & 41.00 & - & 10.80 & - & ICCV 2019\\
RLDA~\cite{RLDA} & F-RCNN & I & 31.10 & 42.98 & 68.10 & 11.88 & 32.11 & ICCV 2019\\
PDA~\cite{PDA} & F-RCNN & V & 30.20 & 43.90 & 55.80 & 13.70 & 53.52 & WACV 2020\\
SimROD (self-adapt) & YOLOv5 & S416 & 31.61 & 35.94 & 56.15 & 4.33 & 17.65 & Ours\\
SimROD (w. teacher X640) & YOLOv5 & S416 & 31.61 & 43.55 & 56.15 & 11.94 & 48.66 & Ours\\
SimROD (w. teacher X1280) & YOLOv5 & S416 & \bftab 31.61 & \bftab 45.66 & 56.15 & \bftab 14.05 & \bftab 57.27 & Ours\\
\midrule
SCDA~\cite{SCDA} & F-RCNN & V & \bftab 37.40 & 42.60 & - & 5.20 & - & CVPR 2019\\
EveryPixelMatters~\cite{Every} & FCOS & R & 35.30 & 45.00 & 70.40 & 9.70 & 27.64 & ECCV 2020\\
SimROD (self adapt) & YOLOv5 & M416 & 36.09 & 42.94 & 59.29 & 6.85 & 29.51 & Ours\\
SimROD (w. teacher X640) & YOLOv5 & M416 & 36.09 & 45.29 & 59.29 & 9.19 & 39.64 & Ours\\
SimROD (w. teacher X1280) & YOLOv5 & M416 & 36.09 & \bftab 47.52 & 59.29 & \bftab 11.43 & \bftab 49.26 & Ours\\
\midrule
SimROD (self-adapt) & YOLOv5 & X640 & 45.67 & 50.81 & 72.18 & 5.14 & 19.38 & Ours\\
SimROD (self-adapt) & YOLOv5 & X1280 & 52.07 & 58.25 & 82.50 & 6.18 & 20.31 & Ours\\
\bottomrule
\end{tabular}
\vspace{2pt}
\caption{Results of different method/model pairs on the KITTI to Cityscapes adaptation scenario. $\tau$ and $\rho$ are the absolute gain and the effective gain respectively as defined in (\ref{tau}) and (\ref{eq:rho}).}
\label{supp:kitti_table}
\end{table*}

\subsection{Qualitative visualization}

\begin{figure*}
\centering\includegraphics[width=2.2\columnwidth]{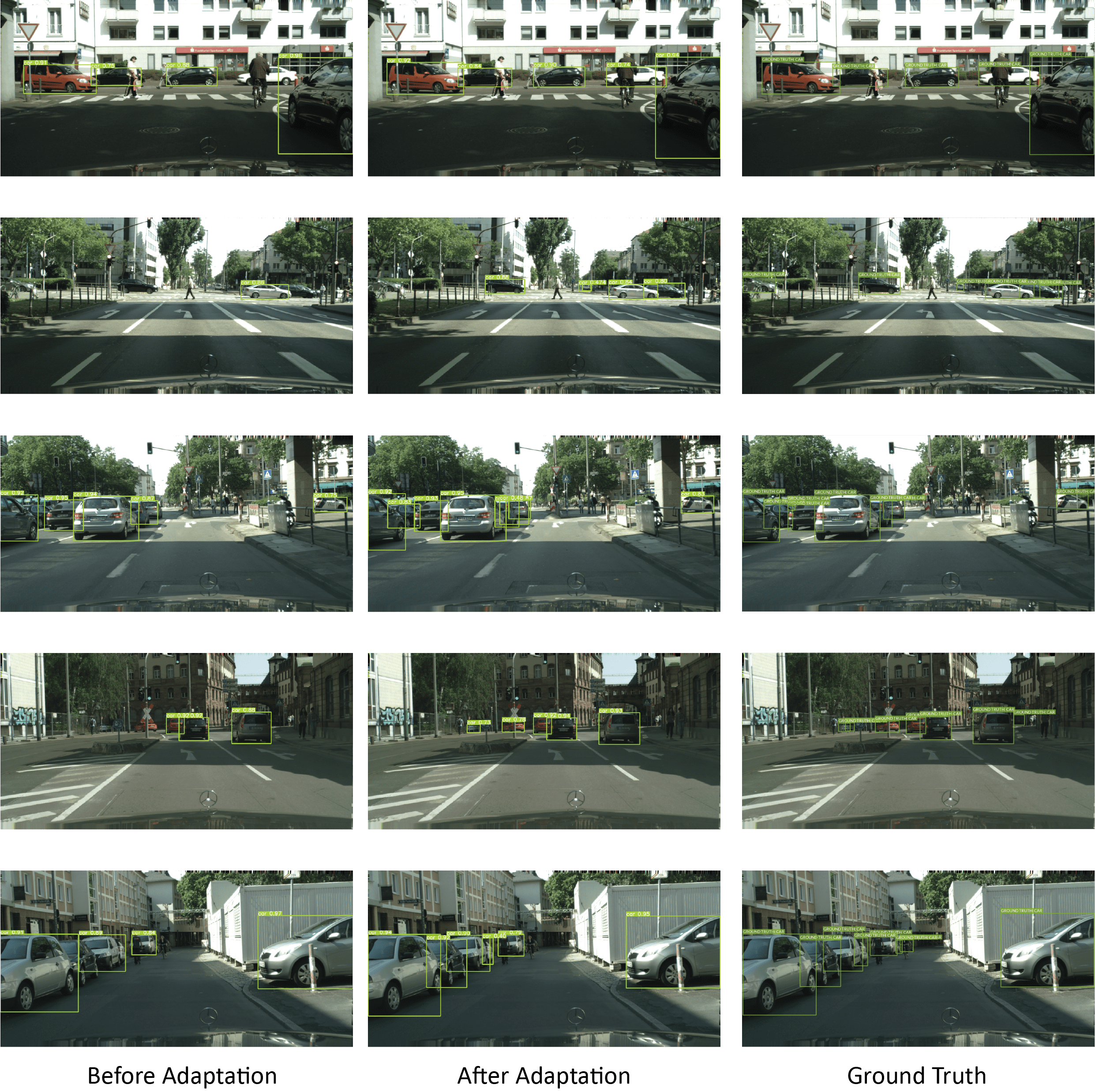}
\vspace{-10pt}
\caption{\small Examples of prediction results on Sim10K to Cityscapes. We show predictions on the target test set before and after applying SimROD as well as the ground-truth labels.}
\label{fig:qual-city}
\end{figure*}

In Figure \ref{fig:qual-city}, we present qualitative results for the detection of the model S416 (i.e. yolov5s with input 416) to demonstrate the improvement brought by SimROD compared to the source model. By comparing with ground-truth labels, Figure \ref{fig:qual-city} shows that the adapted model can detect most objects with good accuracy except for some highly occluded ones. 

\section{More results on artistic benchmark}

\subsection{Benchmark results on Clipart and Comic}

We include the benchmarks results for Clipart and Comic in Table \ref{tab:clipart-full} and \ref{tab:comic-full} respectively. We used only 500 unlabeled images from the target domain for Clipart and 1000 images for Comic. Similar to the results for Watercolor in Table \ref{tab:water}, our method SimROD outperformed the baselines when compared with models that achieve same Source AP performance. Compared to DT+PL in \cite{cross-domain}, our method further improved the AP50 of the S416 model by absolute 8.35, 12 and 10.69 percentage points on Clipart, Comic and Watercolor respectively. In addition, SimROD consistently achieves high effective adaptation gains $\rho$ between 70-97\% across model sizes and benchmarks. 

\begin{table*}
\centering
\begin{tabular}{lllrrrrrl}
\toprule
Method & Arch. & Backbone & Source & AP50 & Oracle & $\tau$ & $\rho$ & Reference\\
\midrule
ADDA~\cite{ADDA} & SSD & V & 26.80 & 27.40 & 55.40 & 0.60 & 2.10 & CVPR 2017\\
DT+PL~\cite{cross-domain} & SSD & V & \bftab 26.80 & \bftab 46.00 & 55.40 & \bftab 19.20 & \bftab 67.13 & CVPR 2018\\
\midrule
DAF~\cite{DAF} & F-RCNN & V & 26.20 & 22.40 & 50.00 & -3.80 & -15.97 & CVPR 2018\\
DT+PL~\cite{cross-domain} & F-RCNN & V & 26.20 & 34.90 & 50.00 & 8.70 & 36.55 & CVPR 2018\\
SWDA~\cite{SWDA} & F-RCNN & V & \bftab 27.80 & 38.10 & 50.00 & 10.30 & 46.40 & CVPR 2019\\
DAM~\cite{DAM} & F-RCNN & V & 24.90 & 41.80 & 50.00 & \bftab 16.90 & \bftab 67.33 & CVPR 2018\\
\midrule
DeepAugment~\cite{DeepAugment} & YOLOv5 & S416 & 29.32 & 31.65 & 56.07 & 2.33 & 8.71 & arXiv 2020\\
{BN-Adapt}~\cite{bn} & YOLOv5 & S416 & 29.32 & 37.43 & 56.07 & 8.11 & 30.32 & NeurIPS 2020\\
Stylize~\cite{Geirhos19} & YOLOv5 & S416 & 29.32 & 38.80 & 56.07 & 9.48 & 35.44 & arXiv 2019\\
STAC~\cite{STAC} & YOLOv5 & S416 & 29.32 & 39.64 & 56.07 & 10.32 & 38.58 & arXiv 2020\\
DT+PL~\cite{cross-domain} & YOLOv5 & S416 & 29.32 & 39.49 & 56.07 & 10.17 & 38.02 & CVPR 2018\\
SimROD (self-adapt) & YOLOv5 & S416 & 29.32 & 41.28 & 56.07 & 11.96 & 44.72 & Ours\\
SimROD (teacher X416) & YOLOv5 & S416 & \bftab29.32 & \bftab47.84 & 56.07 & \bftab 18.52 & \bftab 69.24 & Ours\\
\bottomrule
\end{tabular}
\vspace{2pt}
\caption{\label{tab:clipart-full} Benchmark results on Real (VOC) to Clipart1k domain shift}
\end{table*}

\begin{table*}
\centering
\begin{tabular}[t]{lllrrrrrl}
\toprule
Method & Arch. & Backbone & Source & AP50 & Oracle & $\tau$ & $\rho$ & Reference\\
\midrule
ADDA & SSD & V & 24.90 & 23.80 & 46.40 & -1.10 & -5.12 & CVPR 2017\\
DT & SSD & V & 24.90 & 29.80 & 46.40 & 4.90 & 22.79 & CVPR 2018\\
DT+PL & SSD & V & \bftab24.90 & \bftab37.20 & 46.40 & \bftab12.30 & \bftab57.21 & CVPR 2018\\
\midrule
DAF & F-RCNN & V & 21.40 & 23.20 &  -  & 1.80 &  -  & CVPR 2018\\
DT & F-RCNN & V & 21.40 & 29.80 &  -  & 8.40 &  -  & CVPR 2018\\
SWDA & F-RCNN & V & 21.40 & 28.40 &  -  & 7.00 &  -  & CVPR 2019\\
DAM & F-RCNN & V & \bftab21.40 & \bftab34.50 &  -  & \bftab13.10 &  -  & CVPR 2019\\
\midrule
DeepAugment & YOLOv5 & S416 & 18.19 & 21.39 & 39.81 & 3.20 & 14.80 & arXiv 2020\\
BN-Adapt & YOLOv5 & S416 & 18.19 & 25.53 & 39.81 & 7.34 & 33.95 & NeurIPS 2020\\
Stylize & YOLOv5 & S416 & 18.19 & 27.57 & 39.81 & 9.38 & 43.39 & arXiv 2019\\
STAC & YOLOv5 & S416 & 18.19 & 26.40 & 39.81 & 8.21 & 37.97 & arXiv 2020\\
DT+PL & YOLOv5 & S416 & 18.19 & 25.66 & 39.81 & 7.47 & 34.55 & CVPR 2018\\
SimROD (self-adapt) & YOLOv5 & S416 & 18.19 & 29.54 & 39.81 & 11.35 & 52.50 & Ours\\
SimROD (teacher X416) & YOLOv5 & S416 & \bftab18.19 & \bftab37.65 & 39.81 & \bftab19.46 & \bftab90.01 & Ours\\
\midrule
DeepAugment & YOLOv5 & M416 & 23.58 & 27.65 & 49.13 & 4.07 & 15.93 & arXiv 2020\\
BN-Adapt & YOLOv5 & M416 & 23.58 & 32.04 & 49.13 & 8.46 & 33.11 & NeurIPS 2020\\
Stylize & YOLOv5 & M416 & 23.58 & 34.56 & 49.13 & 10.98 & 42.97 & arXiv 2019\\
STAC & YOLOv5 & M416 & 23.58 & 32.76 & 49.13 & 9.18 & 35.93 & arXiv 2020\\
DT+PL & YOLOv5 & M416 & 23.58 & 33.53 & 49.13 & 9.95 & 38.94 & CVPR 2018\\
SimROD (self-adapt) & YOLOv5 & M416 & 23.58 & 37.93 & 49.13 & 14.35 & 56.15 & Ours\\
SimROD (teacher X416) & YOLOv5 & M416 & \bftab23.58 & \bftab42.08 & 49.13 & \bftab18.50 & \bftab72.41 & Ours\\
\bottomrule
\end{tabular}
\vspace{2pt}
\caption{\label{tab:comic-full} Benchmark results on Real (VOC) to Comic domain shift}
\end{table*}

\subsection{Data efficiency analysis on Watercolor and Comic}

\begin{figure*}
\centering\includegraphics[width=0.9\columnwidth]{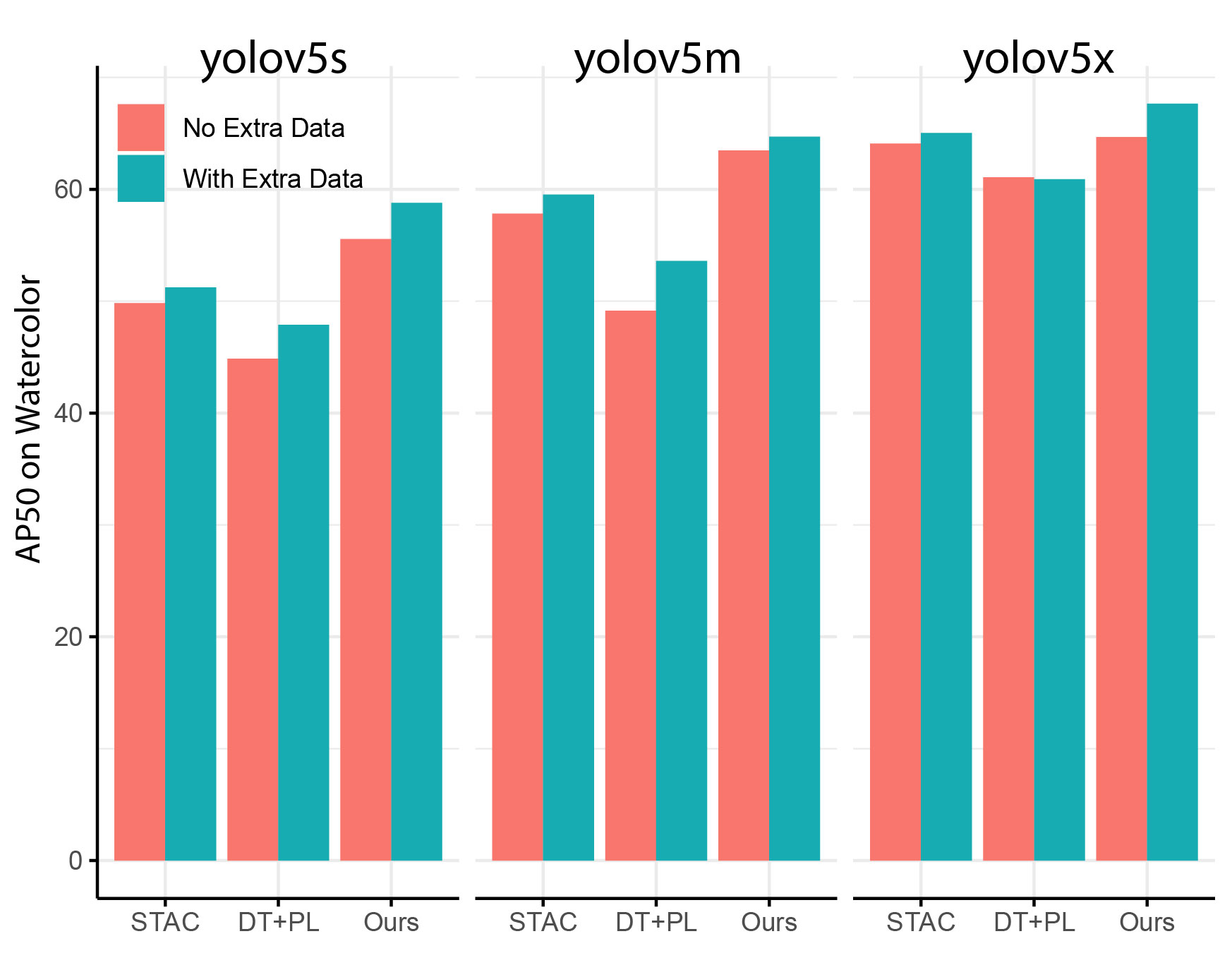}\vspace{-8pt}
\caption{\small Comparison of performance with and without extra unlabeled data on Watercolor.}
\label{water-extra-bar}
\end{figure*}

\begin{figure*}
\centering\includegraphics[width=0.8\columnwidth]{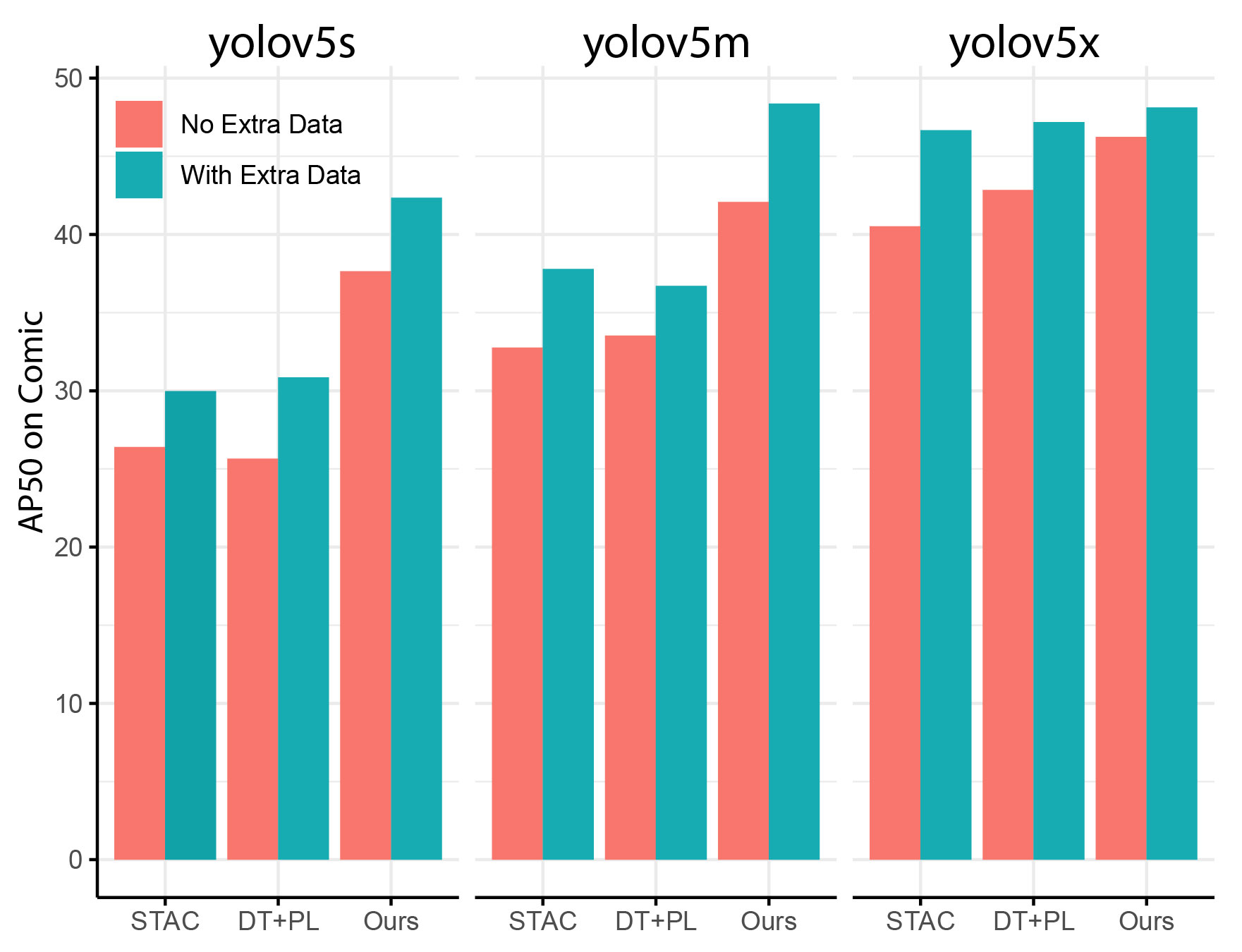}\vspace{-8pt}
\caption{\small Performance comparison on Comic with and without extra unlabeled data.}
\label{comicwater-extra-bar}
\end{figure*}

Next, we analyze the data efficiency of SimROD by increasing the size of unlabeled data used to adapt the models. For Watercolor and Comic, we used the extra splits, which contains extra 52.8K and 17.8K additional unlabeled images respectively. Moreover, all models use the same input size of 416. Figure \ref{water-extra-bar} and \ref{comicwater-extra-bar}  compare the performance of SimROD with the two pseudo-labeling baselines (STAC and DT+PL) on Watercolor and Comic respectively. All methods improved when using more unlabeled data from the target domain. For example, SimROD improves the Yolov5s model performance by absolute +3.23\% and +4.69\% on Watercolor and Comic respectively.

Nonetheless, SimROD could outperform baseline methods without using extra data for Yolov5s and Yolov5m models, which are adapted using the self-adapted teacher Yolov5x. In other words, our proposed method used only 1000 unlabeled images and still outperformed the baselines, which used 50$\times$ or 18$\times$ more data. For example, our method achieved an AP50 of 42.34\% on yolov5s whereas the best baseline on yolov5m has an AP50 of only 37.79\%.

\subsection{Qualitative comparison on Clipart, Comic and Watercolor}

\begin{figure*}
 \centering\includegraphics[width=2.0\columnwidth]{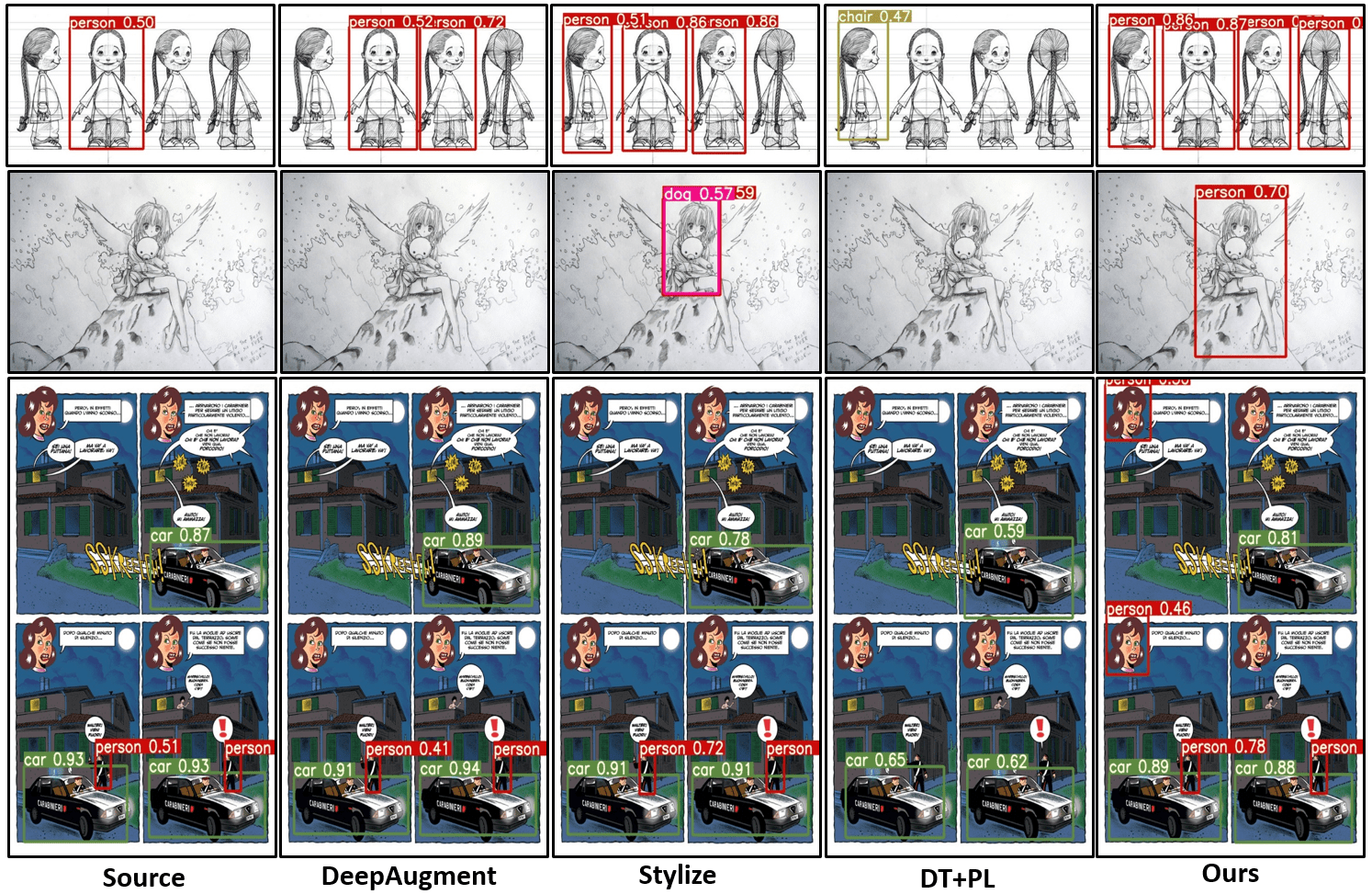}\vspace{-6pt}
 \caption{\small Comparing various methods on examples from the Comic dataset.}
 \label{supp:comic_more_results}
 \end{figure*}

\begin{figure*}
 \centering\includegraphics[width=2.0\columnwidth]{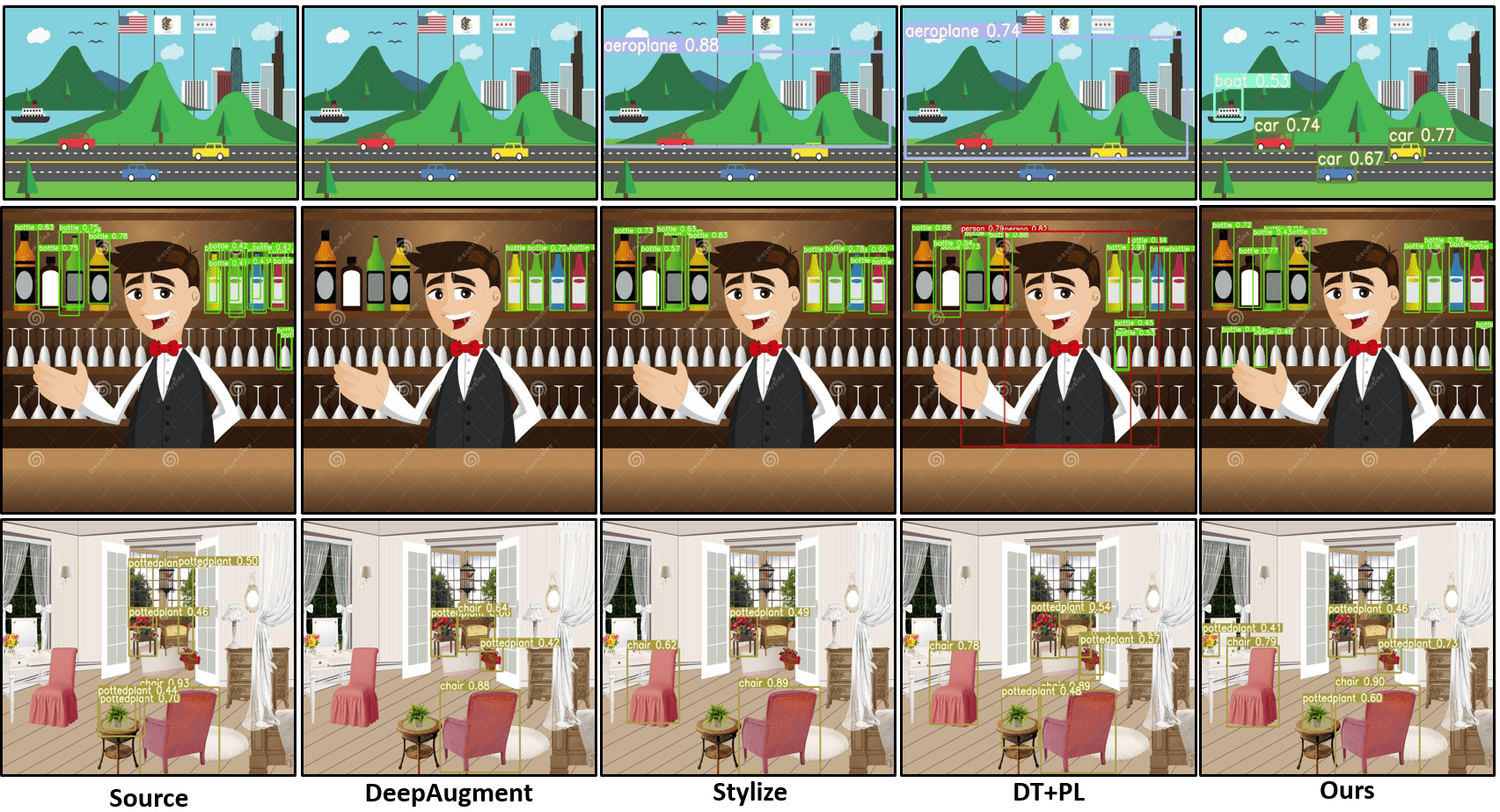}\vspace{-6pt}
 \caption{\small Comparing various methods on examples from the Clipart dataset.}
 \label{supp:clipart_more_results}
 \end{figure*}

In Figures \ref{supp:comic_more_results} and \ref{supp:clipart_more_results}, we provide qualitative comparisons with pseudo-labeling baselines (STAC \cite{STAC} and DT+PL \cite{cross-domain}) and DeepAugment method using same Yolov5s model. These comparisons illustrates the simplicity and effectiveness of SimROD. Our proposed DomainMix augmentation and teacher-guided gradual adaptation enabled to leverage unlabeled target data and to mitigate the label noise and domain shift. In contrast to DT+PL, SimROD did not need to generate synthetic intermediate dataset and our proposed augmentation is much simpler than DeepAugment.

\section{More results on image corruptions}

\subsection{Results for different model sizes}
\label{sup:different-sizes}
Table \ref{tab:result-pascal-c}, \ref{tab:result-coco-c}, and \ref{tab:result-city-c} show only the results for Yolov5m model for Pascal-C, COCO-C, and Cityscapes-C respectively. In Table \ref{sup:tab:result-pascal-c}, \ref{sup:tab:result-coco-c}, and \ref{sup:tab:city-c}, we show that SimROD consistently achieves higher performance compared to the baselines across different model sizes and benchmarks. As expected, larger models provided extra capacity and thus higher Performance.

\begin{table*}
\centering
\fontsize{8}{10}\selectfont
\begin{tabular}{p{2.5cm}p{1cm}p{1cm}ll}
\toprule
Method & $\text{AP}^{50}_{\text{clean}}$ & $\text{mPC}^{50}$ & $\text{rPC}$ & $\tau_c$ \\
\midrule
\addlinespace[0.3em]
\multicolumn{4}{l}{\hspace{-0.5em} \textbf{yolov5s}}\\
\hspace{1em}Source & 75.87 & 42.38 & 55.86 & 0.00\\
\hspace{1em}Stylize & 77.26 & 52.12 & 67.46 & 9.74\\
\hspace{1em}BN-Adapt & 74.71 & 53.75 & 71.94 & 11.37\\
\hspace{1em}DeepAugment & 77.89 & 55.42 & 71.15 & 13.04\\
\hspace{1em}STAC & \bftab  80.11 & 56.12 & 70.05 & 13.74\\
\hspace{1em}\bftab SimROD (Ours) & 80.08 & \bftab 67.95 & \bftab 84.85 & \bftab  25.57\\\cmidrule{2-5}
\hspace{1em}Supervised training & 80.44 & 71.18 & 88.49 & 28.80\\
\addlinespace[0.3em]
\multicolumn{4}{l}{\hspace{-0.5em} \textbf{yolov5m}}\\
\hspace{1em}Source & 83.13 & 53.78 & 64.69 & 0.00\\
\hspace{1em}Stylize & 84.79 & 62.92 & 74.21 & 9.14\\
\hspace{1em}BN-Adapt & 83.01 & 64.60 & 77.82 & 10.82\\
\hspace{1em}DeepAugment & 85.05 & 64.88 & 76.28 & 11.10\\
\hspace{1em}STAC & \bftab 87.00 & 66.88 & 76.88 & 13.11\\
\hspace{1em}\bftab SimROD (Ours) & 86.97 & \bftab 75.40 & \bftab 86.70 & \bftab 21.63\\\cmidrule{2-5}
\hspace{1em}Supervised training & 86.75 & 78.74 & 90.76 & 24.96\\
\addlinespace[0.3em]
\multicolumn{4}{l}{\hspace{-0.5em} \textbf{yolov5x}}\\
\hspace{1em}Source & 87.42 & 62.84  & 71.88 & 0.00\\
\hspace{1em}Stylize & 87.29 & 69.60 & 79.73  & 6.76\\
\hspace{1em}BN-Adapt & 86.59 & 71.59 & 82.68 & 8.75\\
\hspace{1em}DeepAugment & 87.78 & 72.15 & 82.19 & 9.31\\
\hspace{1em}STAC & \bftab 89.57 & 73.68 & 82.25 & 10.84\\
\hspace{1em}\bftab SimROD (Ours) & 89.24 & \bftab 78.48 & \bftab 87.95 & \bftab 15.64\\\cmidrule{2-5}
\hspace{1em}Supervised training & 88.88 & 82.56 & 92.89 & 19.72\\
\bottomrule
\end{tabular}
\vspace{-0pt}
\caption{\label{sup:tab:result-pascal-c}\small Performance comparison on Pascal-C benchmark} 
\end{table*}

\begin{table*}
\centering
\fontsize{8}{10}\selectfont
\begin{tabular}{lllll}
\toprule
Method & $\text{AP}_{\text{clean}}$ & $\text{mPC}$ & $\text{rPC}$ & $\tau_c$ \\
\midrule
\addlinespace[0.3em]
\multicolumn{4}{l}{\hspace{-0.5em} \textbf{yolov5s}}\\
\hspace{1em}Source & \bftab 31.35 & 17.68 & 56.40 & 0.00\\
\hspace{1em}Stylize & 30.07 & 18.99 & 63.15 & 1.31\\
\hspace{1em}BN-Adapt & 30.91 & 20.09 & 64.99 & 2.40\\
\hspace{1em}DeepAugment & 30.37 & 19.87 & 65.44 & 2.19\\
\hspace{1em}STAC & 31.25 & 20.00 & 64.02 & 2.32 \\
\hspace{1em}\bftab SimROD (Ours) & 31.21 & \bftab 23.94 & \bftab 76.71  & \bftab 6.26\\\cmidrule{2-5}
\hspace{1em}Supervised training & 30.90 & 25.33 & 81.99 & 7.65\\
\addlinespace[0.3em]
\multicolumn{4}{l}{\hspace{-0.5em} \textbf{yolov5m}}\\
\hspace{1em}Source & \bftab 36.85 & 22.03 & 59.79 & 0.00\\
\hspace{1em}Stylize & 35.75 & 23.82 & 66.63 & 1.79\\
\hspace{1em}BN-Adapt & 36.24 & 24.79 & 68.39 & 2.76\\
\hspace{1em}DeepAugment & 35.51 & 24.33 & 68.52 & 2.30\\
\hspace{1em}STAC & 36.76 & 24.80 & 67.46 & 2.77\\
\hspace{1em}\bftab SimROD (Ours) & 36.79 & \bftab 28.46 & \bftab 77.36  & \bftab 6.43\\\cmidrule{2-5}
\hspace{1em}Supervised training & 36.23 & 30.16 & 83.26 & 8.13\\
\addlinespace[0.3em]
\multicolumn{4}{l}{\hspace{-0.5em} \textbf{yolov5x}}\\
\hspace{1em}Source & 41.61 & 26.60 & 63.93 & 0.00\\
\hspace{1em}Stylize & 40.38 & 28.16 & 69.73 & 1.56\\
\hspace{1em}BN-Adapt & 41.70 & 29.77 & 71.40 & 3.17\\
\hspace{1em}DeepAugment & 41.12 & 29.13 & 70.84 & 2.53\\
\hspace{1em}STAC & \bftab 41.85 & 29.69 & 70.93 & 3.09\\
\hspace{1em}\bftab SimROD (Ours) & 41.63 & \bftab 31.87 & \bftab 76.57 & \bftab 5.27\\\cmidrule{2-5}
\hspace{1em}Supervised training & 41.06 & 34.84 & 84.86 & 8.24\\
\bottomrule
\vspace{-8pt}
\end{tabular}
\caption{\label{sup:tab:result-coco-c}\small Performance benchmark on COCO-C dataset}
\end{table*}

\begin{table*}
\centering
\fontsize{8}{10}\selectfont
\begin{tabular}{p{2.5cm}rrrr}
\toprule
Method & $\text{AP}_{\text{clean}}$ & $\text{mPC}$ & $\text{rPC}$ & $\tau_c$ \\
\midrule
\addlinespace[0.3em]
\multicolumn{5}{l}{\hspace{-0.5em} \textbf{yolov5s}}\\
\hspace{1em}Source & 17.08 & 9.50 & 55.62 & 0.00\\
\hspace{1em}Stylize & 18.96 & 11.75 & 61.97 & 2.25\\
\hspace{1em}DeepAugment & 17.24 & 11.39 & 66.07 & 1.89\\
\hspace{1em}STAC & \bftab 20.34 & 12.82 & 63.02 & 3.32\\
\hspace{1em}\bftab SimROD (Ours) & 19.82 & \bftab 14.95 & \bftab 75.45 & \bftab 5.45\\\cmidrule{2-5}
\hspace{1em}Supervised training & 22.30 & 19.35 & 86.77 & 9.85\\
\addlinespace[0.3em]
\multicolumn{5}{l}{\hspace{-0.5em} \textbf{yolov5m}}\\
\hspace{1em}Source & 19.48 & 11.53 & 59.19 & 0.00\\
\hspace{1em}Stylize & 21.77 & 14.62 & 67.16 & 3.09\\
\hspace{1em}DeepAugment & 20.28 & 14.79 & 72.93 & 3.26\\
\hspace{1em}STAC & \bftab 24.54 & 15.39 & 62.71 & 3.86\\
\hspace{1em}\bftab SimROD (Ours) & 24.06 & \bftab 18.01 & \bftab 74.86 & \bftab 6.48\\\cmidrule{2-5}
\hspace{1em}Supervised training & 26.58 & 23.50 & 88.43 & 11.97\\
\addlinespace[0.3em]
\multicolumn{5}{l}{\hspace{-0.5em} \textbf{yolov5x}}\\
\hspace{1em}Source & 25.65 & 16.63 & 64.83 & 0.00\\
\hspace{1em}Stylize & 27.70 & 19.38 & 69.96 & 2.75\\
\hspace{1em}DeepAugment & 25.12 & 18.80 & \bftab 74.84 & 2.17\\
\hspace{1em}STAC & \bftab 29.62 & 20.98 & 70.85 & 4.35\\
\hspace{1em}\bftab SimROD (Ours) & 29.27 & \bftab 21.70 & 74.15 & \bftab 5.07\\\cmidrule{2-5}
\hspace{1em}Supervised training & 31.48 & 27.66 & 87.87 & 11.03\\
\bottomrule
\end{tabular}
\vspace{-0pt}
\caption{\label{sup:tab:city-c} Performance benchmark on Cityscapes-C dataset}
\end{table*}

\subsection{Per-corruption performance on Pascal-C}\label{per-corruption-result}

In the main paper, we reported the mAP, rPC, and $\tau_c$ metrics, which were averaged over 15 corruption types. Here, in Tables \ref{tab:per-cor-5s}, \ref{tab:per-cor-5m}, and \ref{tab:per-cor-5x}, we provide a breakdown of the results for each corruption type on the Pascal-C dataset for the three YOLOv5 models.


\begin{table*}[ht]
\centering
\fontsize{9}{11}\selectfont
\resizebox{\textwidth}{!}{\begin{tabular}{p{2cm}lllllllllllllllll}
\toprule
\multicolumn{3}{c}{ } & \multicolumn{3}{c}{Noise} & \multicolumn{4}{c}{Blur} & \multicolumn{4}{c}{Weather} & \multicolumn{4}{c}{Digital} \\
\cmidrule(l{3pt}r{3pt}){4-6} \cmidrule(l{3pt}r{3pt}){7-10} \cmidrule(l{3pt}r{3pt}){11-14} \cmidrule(l{3pt}r{3pt}){15-18}
Method & $\text{AP}_{\text{clean}}$ & $\text{mPC}$ & Gauss. & Shot & Impulse & Defocus & Glass & Motion & Zoom & Snow & Frost & Fog & Bright & Contrast & Elastic & Pixel & JPEG\\
\midrule
Source & 75.87 & 42.38 & 32.71 & 35.32 & 28.24 & 43.02 & 32.96 & 39.87 & 29.05 & 37.09 & 43.53 & 59.66 & 69.21 & 42.00 & 47.04 & 46.53 & 49.48\\
Stylize & 77.26 & 52.12 & 41.51 & 44.61 & 37.82 & 49.80 & 48.02 & 47.37 & 35.79 & 49.53 & 57.37 & 67.55 & 74.07 & 51.69 & 59.10 & 56.77 & 60.84\\
DeepAugment & 77.89 & 55.42 & 50.48 & 53.12 & 48.67 & 55.38 & 49.23 & 48.87 & 37.58 & 49.73 & 58.19 & 70.29 & 74.91 & 56.88 & 51.61 & 63.39 & 62.99\\
BN Adapt & 74.71 & 53.75 & 48.07 & 51.22 & 46.00 & 53.23 & 44.34 & 48.60 & 38.63 & 50.56 & 55.80 & 68.50 & 73.34 & 57.18 & 59.32 & 52.86 & 58.55\\
STAC & \bftab 80.11 & 56.12 & 46.85 & 49.78 & 44.08 & 58.41 & 45.38 & 51.99 & 41.68 & 53.39 & 59.80 & 74.01 & 78.91 & 59.85 & 61.76 & 56.14 & 59.78\\
\bftab SimROD (Ours) & 80.08 & \bftab 67.95 & \bftab 64.91 & \bftab 66.11 & \bftab 65.28 & \bftab 65.12 & \bftab 63.03 & \bftab 65.54 & \bftab 53.99 & \bftab 69.19 & \bftab 69.27 & \bftab 76.85 & \bftab 79.14 & \bftab 71.38 & \bftab 73.52 & \bftab 65.54 & \bftab 70.34\\
\midrule
Oracle & 80.44 & 71.18 & 68.28 & 69.14 & 68.10 & 68.18 & 67.84 & 69.77 & 62.19 & 71.41 & 71.26 & 77.49 & 79.95 & 73.41 & 75.90 & 70.40 & 74.41\\
\bottomrule
\end{tabular}}
\vspace{-6pt}
\caption{\label{tab:per-cor-5s}Performance comparison per corruption type for YOLOv5s model on Pascal-C benchmark}
\end{table*}


\begin{table*}[ht]
\centering
\fontsize{10}{12}\selectfont
\resizebox{\textwidth}{!}{\begin{tabular}{p{2cm}lllllllllllllllll}
\toprule
\multicolumn{3}{c}{ } & \multicolumn{3}{c}{Noise} & \multicolumn{4}{c}{Blur} & \multicolumn{4}{c}{Weather} & \multicolumn{4}{c}{Digital} \\
\cmidrule(l{3pt}r{3pt}){4-6} \cmidrule(l{3pt}r{3pt}){7-10} \cmidrule(l{3pt}r{3pt}){11-14} \cmidrule(l{3pt}r{3pt}){15-18}
Method & $\text{AP}_{\text{clean}}$ & $\text{mPC}$ & Gauss. & Shot & Impulse & Defocus & Glass & Motion & Zoom & Snow & Frost & Fog & Bright & Contrast & Elastic & Pixel & JPEG\\
\midrule
Source & 83.13 & 53.78 & 47.44 & 51.35 & 44.98 & 53.87 & 42.17 & 48.61 & 36.64 & 51.77 & 56.29 & 71.74 & 78.82 & 55.81 & 56.43 & 54.52 & 56.17\\
Stylize & 84.79 & 62.92 & 53.44 & 57.56 & 52.62 & 60.18 & 57.42 & 57.53 & 45.32 & 63.02 & 67.50 & 78.02 & 81.91 & 65.64 & 69.69 & 66.10 & 67.86\\
DeepAugment & 85.05 & 64.88 & 61.75 & 64.06 & 60.64 & 63.74 & 57.95 & 56.18 & 44.75 & 62.31 & 68.27 & 79.36 & 82.69 & 68.34 & 61.92 & 71.40 & 69.78\\
BN Adapt & 83.01 & 64.60 & 61.06 & 63.83 & 60.54 & 62.33 & 55.29 & 58.77 & 46.71 & 65.44 & 67.88 & 78.34 & 81.62 & 69.48 & 68.81 & 62.15 & 66.75\\
STAC & \bftab 87.00 & 66.88 & 61.46 & 64.77 & 60.73 & 67.17 & 55.54 & 61.35 & 49.57 & 68.41 & 71.20 & 82.52 & 85.90 & 71.83 & 69.92 & 65.61 & 67.25\\
\bftab SimROD (Ours) & 86.97 & \bftab 75.40 & \bftab 72.00 & \bftab 74.11 & \bftab 73.01 & \bftab 72.65 & \bftab 70.25 & \bftab 72.85 & \bftab 60.65 & \bftab 77.81 & \bftab 77.47 & \bftab 84.03 & \bftab 86.17 & \bftab 79.66 & \bftab 80.49 & \bftab 72.54 & \bftab 77.36\\
\midrule
Oracle & 86.75 & 78.74 & 76.35 & 76.68 & 76.42 & 75.63 & 75.12 & 77.10 & 70.31 & 80.07 & 79.56 & 84.25 & 86.15 & 80.60 & 82.88 & 78.73 & 81.22\\
\bottomrule
\end{tabular}}
\vspace{-6pt}
\caption{\label{tab:per-cor-5m}Performance comparison per corruption type for YOLOv5m model on Pascal-C benchmark}
\end{table*}


\begin{table*}[ht]
\centering
\fontsize{10}{12}\selectfont
\resizebox{\textwidth}{!}{\begin{tabular}{p{2cm}lllllllllllllllll}
\toprule
\multicolumn{3}{c}{ } & \multicolumn{3}{c}{Noise} & \multicolumn{4}{c}{Blur} & \multicolumn{4}{c}{Weather} & \multicolumn{4}{c}{Digital} \\
\cmidrule(l{3pt}r{3pt}){4-6} \cmidrule(l{3pt}r{3pt}){7-10} \cmidrule(l{3pt}r{3pt}){11-14} \cmidrule(l{3pt}r{3pt}){15-18}
Method & $\text{AP}_{\text{clean}}$ & $\text{mPC}$ & Gauss. & Shot & Impulse & Defocus & Glass & Motion & Zoom & Snow & Frost & Fog & Bright & Contrast & Elastic & Pixel & JPEG\\
\midrule
Source & 87.42 & 62.84 & 59.30 & 61.06 & 58.38 & 61.45 & 51.43 & 58.48 & 41.79 & 63.82 & 66.34 & 77.28 & 84.25 & 65.77 & 64.40 & 63.07 & 65.79\\
Stylize & 87.29 & 69.60 & 62.44 & 65.03 & 62.20 & 67.57 & 63.64 & 65.13 & 50.84 & 70.44 & 74.10 & 82.44 & 85.30 & 74.16 & 74.73 & 72.76 & 73.25\\
DeepAugment & 87.78 & 72.15 & 71.25 & 73.27 & 71.16 & 71.40 & 64.70 & 65.57 & 49.76 & 71.42 & 74.91 & 84.17 & 86.43 & 77.48 & 68.75 & 77.16 & 74.88\\
BN Adapt & 86.59 & 71.59 & 71.05 & 72.63 & 70.94 & 67.90 & 63.70 & 66.55 & 52.41 & 72.79 & 73.91 & 82.38 & 84.62 & 76.03 & 74.96 & 70.51 & 73.43\\
STAC & \bftab 89.57 & 73.68 & 71.77 & 73.40 & 71.71 & 72.57 & 64.51 & 69.37 & 52.81 & 76.21 & 77.68 & 85.04 & 88.40 & 78.87 & 75.92 & 72.69 & 74.23\\
\bftab SimROD (Ours) & 89.24 & \bftab 78.48 & \bftab 76.09 & \bftab 78.31 & \bftab 77.23 & \bftab 75.85 & \bftab 73.11 & \bftab 75.29 & \bftab 62.75 & \bftab 81.10 & \bftab 80.96 & \bftab 86.62 & \bftab 88.16 & \bftab 82.94 & \bftab 82.45 & \bftab 76.64 & \bftab 79.69\\
\midrule
Oracle & 88.88 & 82.56 & 81.14 & 81.96 & 81.27 & 79.10 & 79.08 & 80.65 & 73.97 & 83.58 & 83.66 & 87.18 & 88.54 & 84.03 & 85.55 & 84.09 & 84.57\\
\bottomrule
\end{tabular}}
\vspace{-6pt}
\caption{\label{tab:per-cor-5x}Performance comparison per corruption type for YOLOv5x model on Pascal-C benchmark}
\end{table*}

\subsection{Performance comparison with Augmix}

Here, we compare our proposed method with Augmix augmentation \cite{AugMix} and report the results on Pascal-C in Table \ref{tab:augmix-5s} and \ref{tab:augmix-5x} for the models YOLOv5s and YOLOv5x respectively. When comparing the mean performance under corruption (mPC), we can see that Augmix performed the worst among all augmentation-based baselines. Interestingly, applying Augmix augmentation with DeepAugment improved the performance of DeepAugment by +3.3\% AP50 and +1.03\% AP50 on YOLOv5s and YOLOv5x models respectively. Nonetheless, SimROD still outperformed DeepAument+Augmix by more than +5\% AP50 on both models. Although we have not tried, it is possible that applying Augmix on top of DomainMix may further improve the performance of our proposed method.
 
\begin{table*}
\centering
\small{
\begin{tabular}{p{4cm}ll}
\toprule
Method & $\text{AP}_{\text{clean}}$ & $\text{mPC}$\\
\midrule
Source & 75.87 & 42.38\\
Augmix & 79.42 & 46.94\\
Stylize & 77.26 & 52.12\\
DeepAugment & 77.89 & 55.42\\
DeepAugment+Augmix & \bftab 80.85 & 60.15\\
\bftab SimROD (Ours) & 80.08 & \bftab 67.95\\
\bottomrule
\end{tabular}}
\vspace{-0pt}
\caption{\label{tab:augmix-5s}Augmix comparison for YOLOv5s model on Pascal-C.}
\end{table*}

\begin{table*}
\centering
\small{
\begin{tabular}{p{4cm}ll}
\toprule
Method & $\text{AP}_{\text{clean}}$ & $\text{mPC}$\\
\midrule
Augmix & 87.46 & 62.31\\
Source & 87.42 & 62.84\\
Stylize & 87.29 & 69.60\\
DeepAugment & 87.78 & 72.15\\
DeepAugment+Augmix & 88.36 & 73.18\\
\bftab SimROD (Ours) & \bftab 89.24 & \bftab 78.48\\
\bottomrule~
\end{tabular}}
\vspace{-0pt}
\caption{\label{tab:augmix-5x}Augmix comparison for YOLOv5x model on Pascal-C.}
\end{table*}

\subsection{Data efficiency analysis on Pascal-C}

\begin{figure*}
\centering\includegraphics[width=1.5\columnwidth]{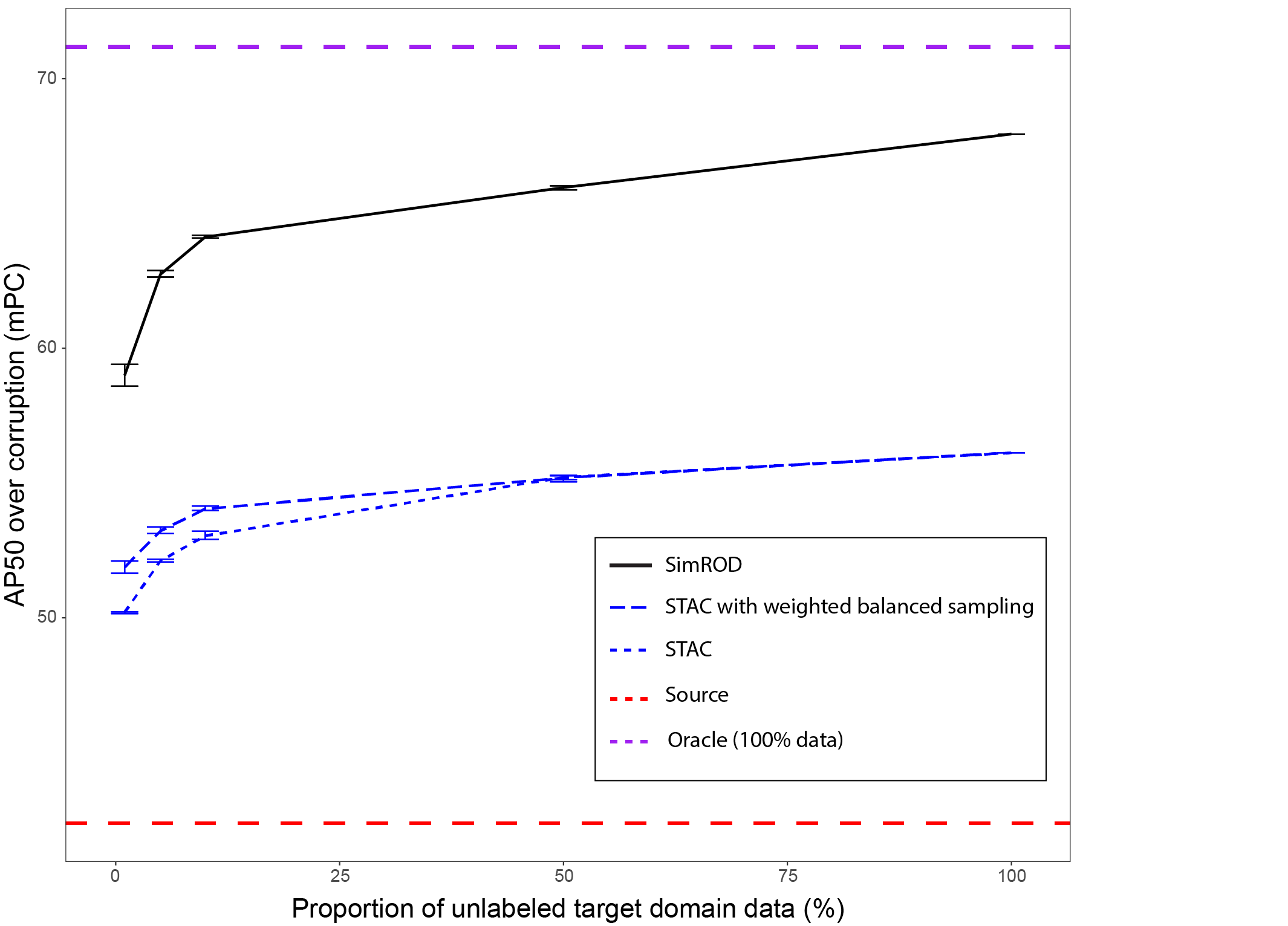}
\vspace{-6pt}
\caption{\small mPC performance of YOLOv5s on Pascal-C for a given percentage of unlabeled target data and using 100\% source data.}
\label{fig:imbalanced}
\end{figure*}

\begin{figure*}
\centering\includegraphics[width=1.5\columnwidth]{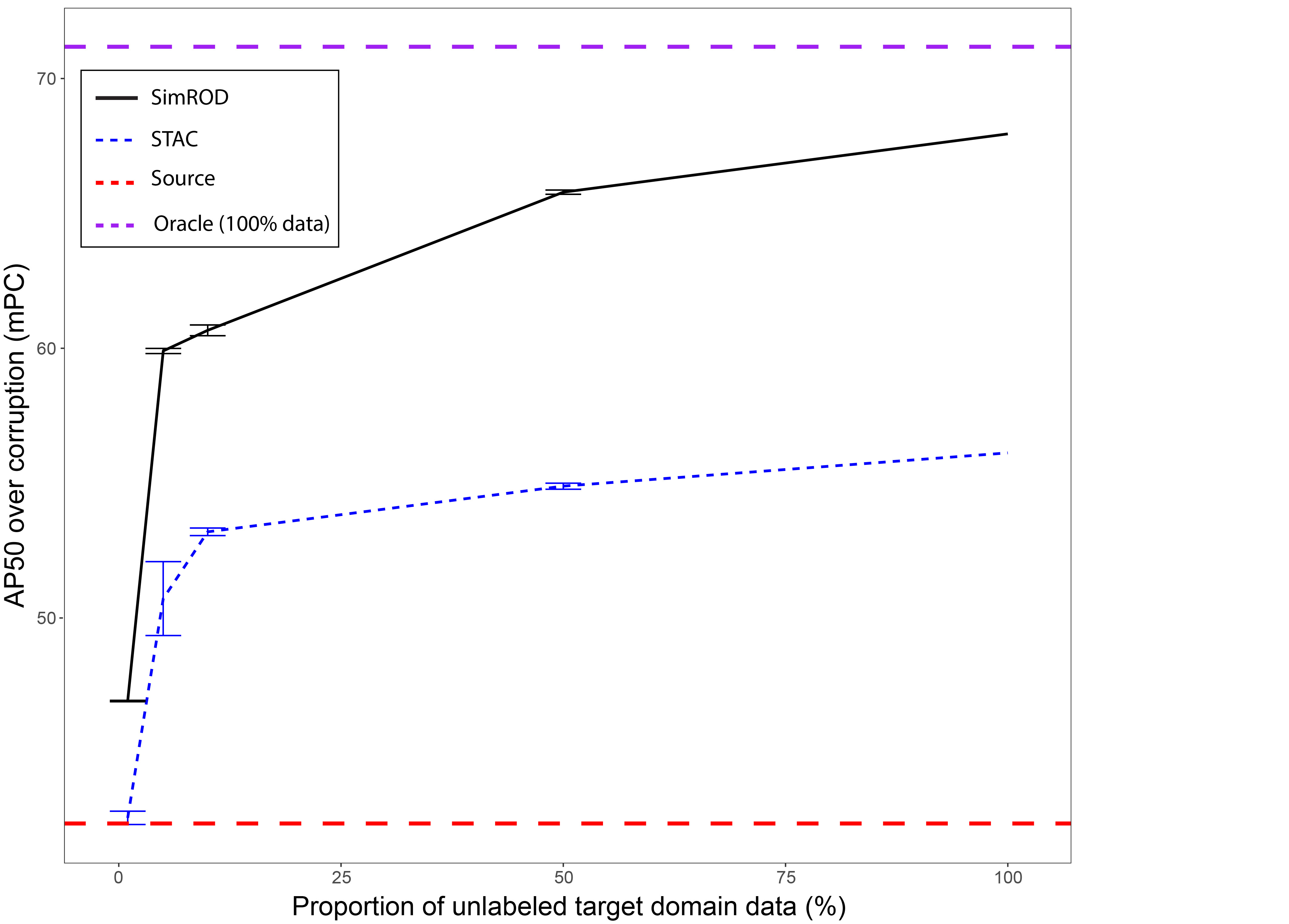}
\vspace{-6pt}
\caption{\small mPC performance of YOLOv5s on Pascal-C for a given percentage of training data (source and target).}
\label{fig:balanced}
\end{figure*}

In Figure \ref{fig:imbalanced} and \ref{fig:balanced}, we analyzed the data efficiency of our proposed method using a YOLOv5s model and Pascal-C dataset. For that, we used a subset of training datasets and considered two scenarios. For both scenarios, we randomly generated three different sets of data, measured the performance in three runs. The average of the three runs are plotted with error bars in Figure \ref{fig:imbalanced} and \ref{fig:balanced}.

In the first scenario, we used all the available labeled data from source domain consisting of 5011 images. On the other hand, we used only a portion of the unlabeled images available. As shown in Figure \ref{fig:imbalanced}, our proposed method outperformed STAC by a margin of 10\% AP50. Moreover, our method achieved a relative robustness $\tau_c$ of +21.75\% AP50 and +16.61\% AP50 using only 10\% and 1\% of unlabeled target domain images respectively. Since the data was imbalanced in this scenario, we also considered applying the weighted balanced sampling to STAC. Figure \ref{fig:imbalanced} shows that it could slightly improve the performance of STAC when the datasets were very imbalanced.

In the second scenario, we used only a given percentage of the available training data for both the source and target domain. While this scenario assumes the datasets are balanced, the total number of training images is much smaller than in the previous scenario. For example, using 1\% of training data corresponds to a total of 165 images. With 1\% of training data, STAC could not adapt the model. In contrast, our proposed method provided a relative robustness $\tau_c$ of +4.54\% AP50 and +18.28\% AP50 using only 1\% and 10\% of training data respectively.

These results confirm that our method was more data-efficient. In particular, our DomainMix augmentation could produce a diverse set of mixed samples even from very few training images from both domains. When more unlabeled data was available, our method could further leverage the unlabeled data and provide strong supervision for adaptation by mitigating the label noise.

\subsection{Effects of corruption severity levels}

\begin{figure*}
\centering\includegraphics[width=1.45\columnwidth]{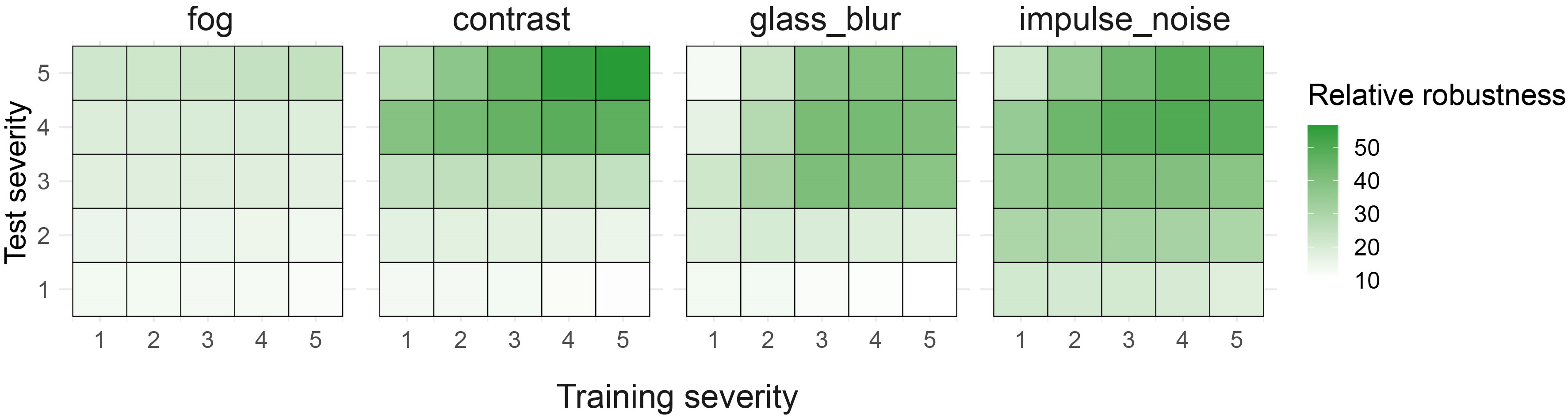}
\vspace{-4pt}
\caption{\small Relative robustness improvement on YOLOv5s using our method for specific corruption types and severity levels on Pascal-C.}
\label{fig:severity_delta}
\end{figure*}

\begin{figure*}
\centering\includegraphics[width=1.4\columnwidth]{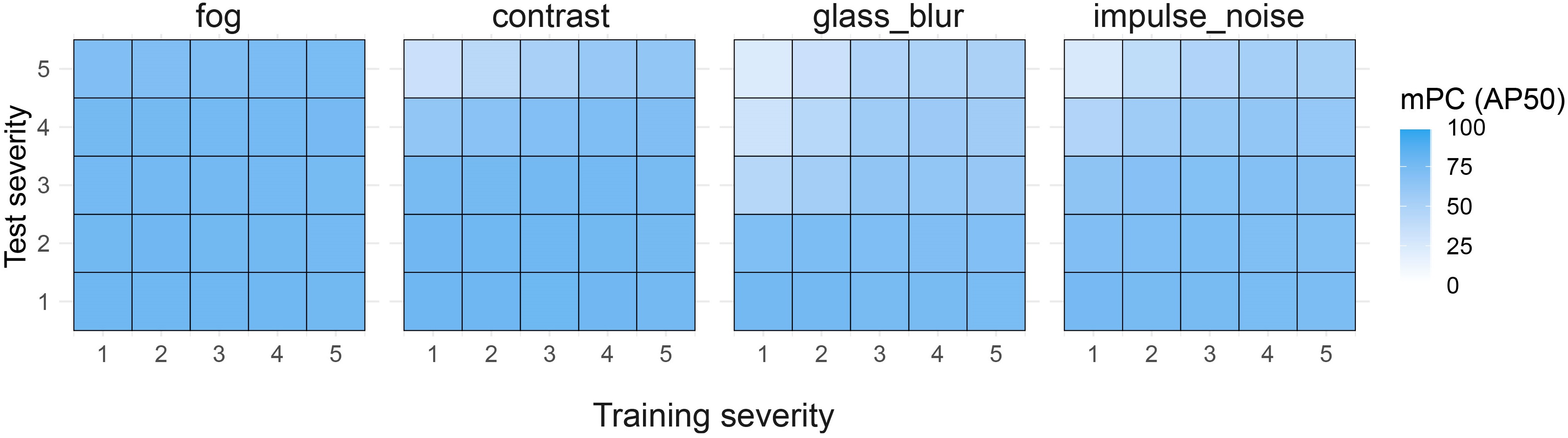}
\vspace{-4pt}
\caption{\small Final mPC performance of YOLOv5s using our method for specific corruption types and severity levels on Pascal-C.}
\label{fig:severity_after}
\end{figure*}

\begin{figure*}
\centering\includegraphics[width=1.45\columnwidth]{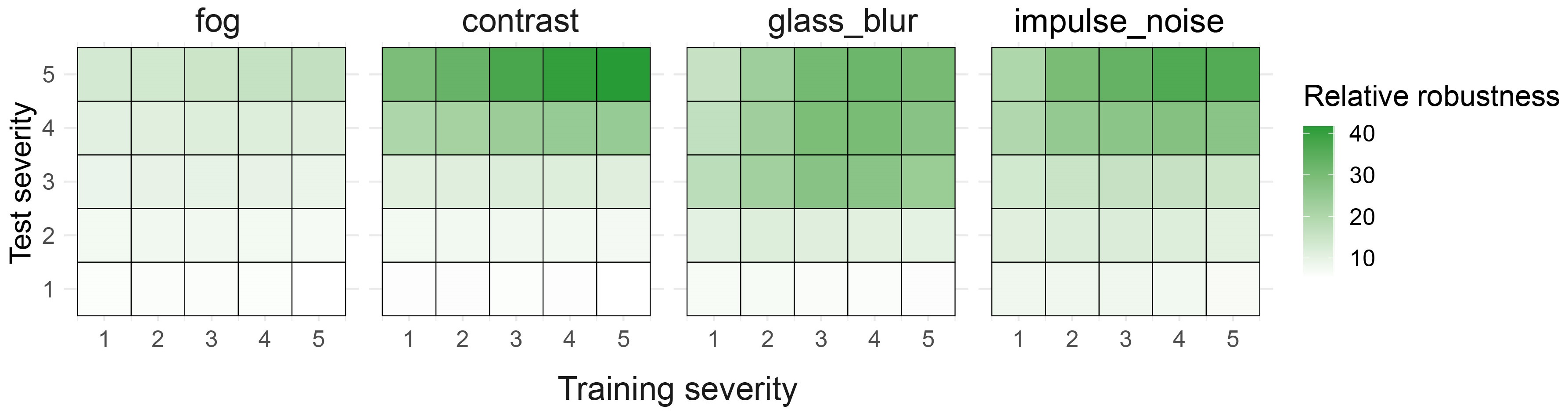}
\vspace{-4pt}
\caption{\small Relative robustness improvement on YOLOv5x using our method for specific corruption types and severity levels on Pascal-C.}
\label{fig:severity_5x_delta}
\end{figure*}

\begin{figure*}
\centering\includegraphics[width=1.4\columnwidth]{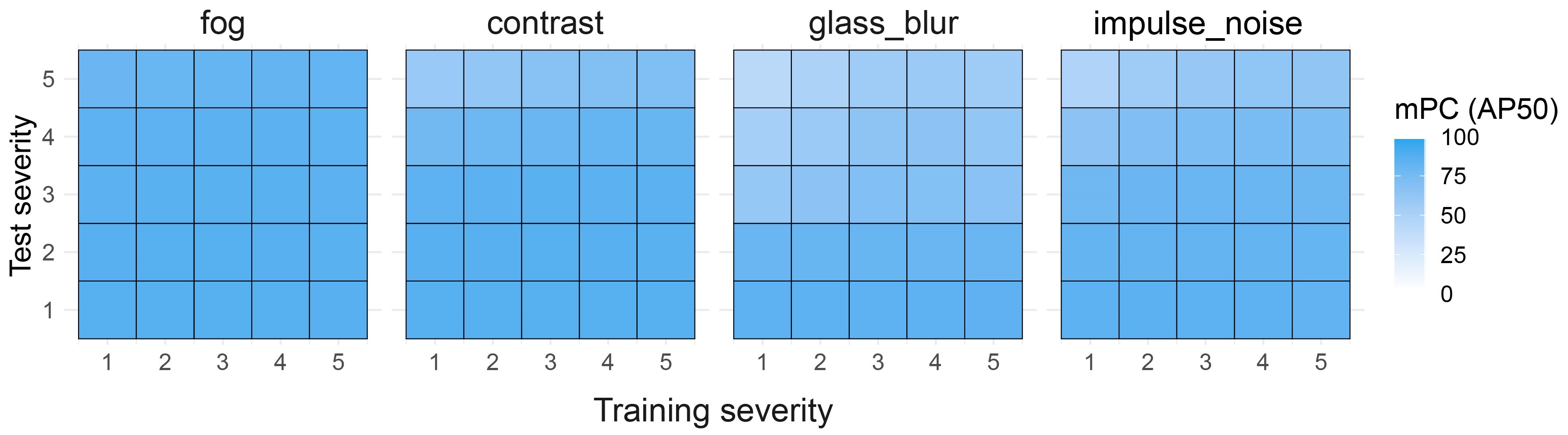}
\vspace{-4pt}
\caption{\small Final mPC performance of YOLOv5x using our method for specific corruption types and severity levels on Pascal-C.}
\label{fig:severity_5x_after}
\end{figure*}

To apply our method on the image corruption benchmark, we applied a corruption severity level of 3 for creating the unlabeled target domain images. In this section, we present additional analysis to understand the effects of corruption severity of the training images on the test performance. 
In Fig. \ref{fig:severity_delta} and \ref{fig:severity_after}, we show the relative robustness $\tau$ and mean performance under corruption $\text{mPC}$ of an adapted Yolov5s model using our method. Similarly, Fig. \ref{fig:severity_5x_delta} and \ref{fig:severity_5x_after} show the same metrics for an adapted YOLOv5x model. 

The corruption types are sorted in ascending order based on the performance of the source model on these types. For instance, the source models achieved the highest mPC on fog and lowest mPC on impulse noise. This explains that the relative robustness on fog  was lower compared to those on other corruption types because the source model already achieved high mPC on fog. Notable improvements were observed on the other corruption types.

Fig. \ref{fig:severity_delta} and \ref{fig:severity_after} show that the adapted YOLOv5s model enjoyed higher improvement on test datasets with higher severity levels. More importantly, high improvements could be achieved when the training images have severity levels similar to those of the test images. This means that using unlabeled target-domain samples is effective as long as they are representative of the actual test set.

\subsection{Qualitative comparison on image corruptions}

\begin{figure*}
 \centering\includegraphics[width=2.0\columnwidth]{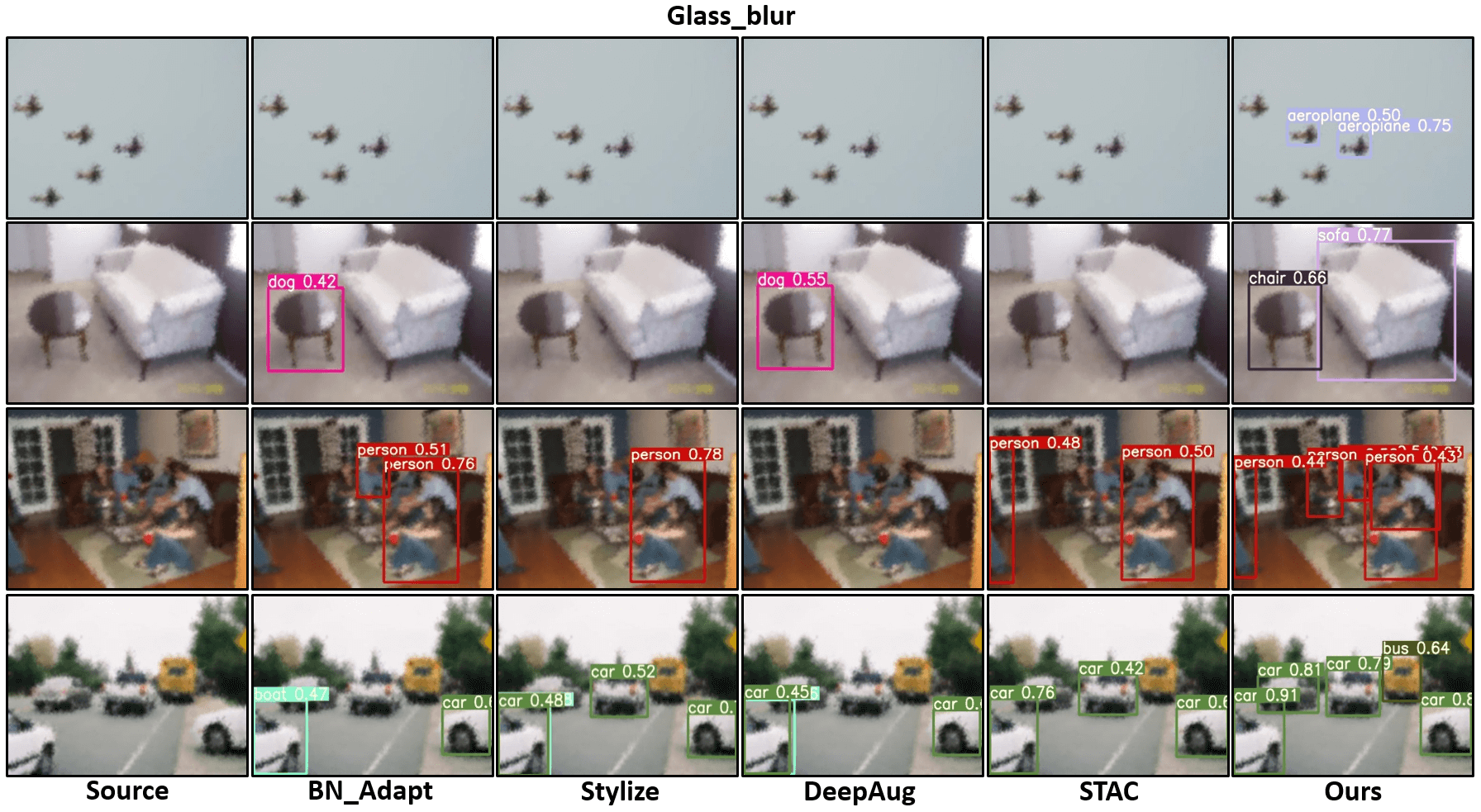}\vspace{-6pt}
 \caption{\small Demonstration of how different methods handle glass\_blur corruption (severity 5); images from Pascal-C.}
 \label{supp:pascal_glass_blur_v2}
 \end{figure*}

\begin{figure*}
\centering\includegraphics[width=2.0\columnwidth]{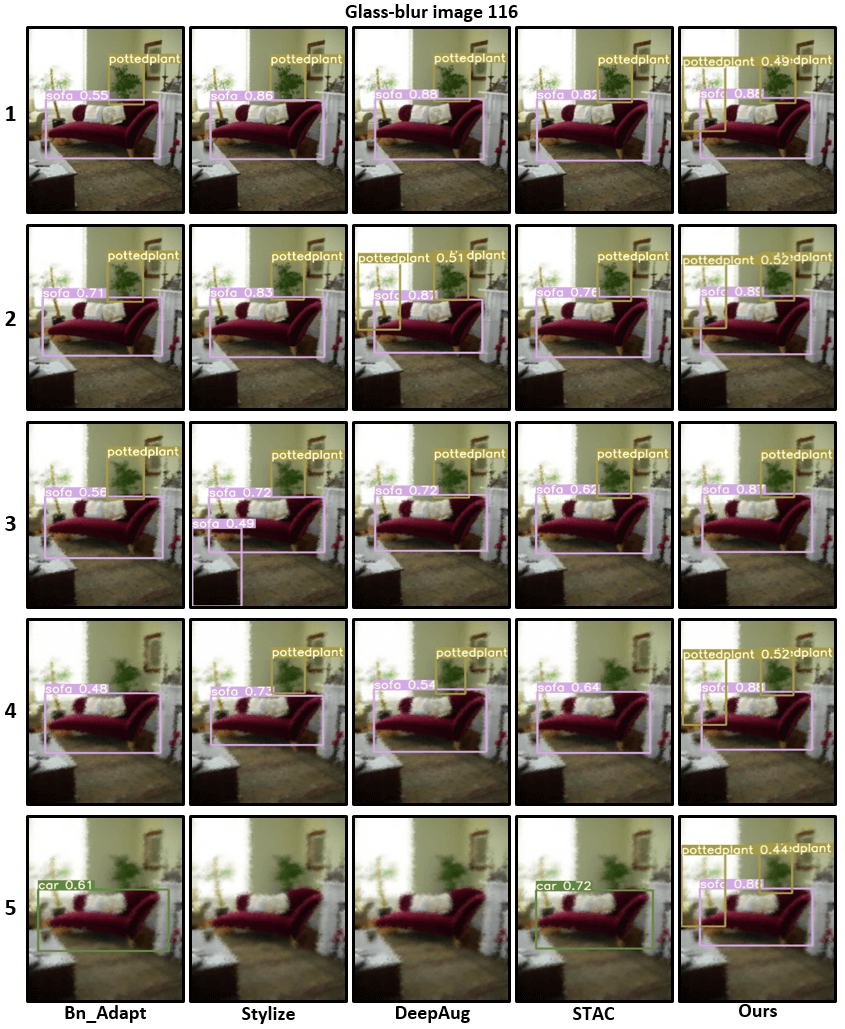}\vspace{-8pt}
\caption{\small Demonstration of how different methods handle glass\_blur corruption at different severity levels; image from Pascal-C.}
\label{supp:pascal_glass_blur_all_severities}
\end{figure*}

Fig. \ref{supp:pascal_glass_blur_v2} illustrates how various methods handle the glass blur corruption (severity 5) on Pascal-C sample. In addition, Fig. \ref{supp:pascal_glass_blur_all_severities} shows results of various methods across a range of severity levels for the glass blur corruption. We see that the proposed method was more effective in handling the corruptions. In contrast to the baseline methods, our adaptation method detected most objects in the images and make fewer classification errors. We could also observe that the source model completely failed to detect objects in most cases. 

\subsection{More detailed ablations on the components}
Table \ref{supp:tab:ablation} expands the ablation study provided in the main paper onto various model sizes.

\begin{table*}
\small\centering
\fontsize{8}{10}\selectfont
\begin{tabular}{p{2.5cm}lccccll}
\toprule
Model & Method & TG & DomainMix & BN-Adapt & Finetune & Corrupt AP50 & $\tau_c$\\
\midrule
\addlinespace[0.3em]
\hspace{1em} & Source &  &  &  &  & 42.38 & 0.00\\
\hspace{1em} & BN-Adapt &  &  & \checkmark &  & 53.75 & 11.37\\
\hspace{1em} & BN-Adapt + DomainMix &  & \checkmark & \checkmark &  & 56.13 & 13.75\\
{\textbf{yolov5s}}\hspace{1em} & SimROD (Ours) w/o Teacher Guidance &  & \checkmark & \checkmark & \checkmark & 60.35 & 17.97\\
\hspace{1em} & SimROD (Ours) w/o Gradual Adaptation & \checkmark & \checkmark &  & \checkmark & 67.87 & 25.49\\
\hspace{1em} & Our full method (SimROD) & \checkmark & \checkmark & \checkmark & \checkmark & 67.95 & 25.57\\
\midrule
\addlinespace[0.3em]
\hspace{1em} & Source &  &  &  &  & 53.78 & 0.00\\
\hspace{1em} & BN-Adapt &  &  & \checkmark &  & 64.60 & 10.82\\
\hspace{1em} & BN-Adapt + DomainMix &  & \checkmark & \checkmark &  & 66.78 & 13.01\\
{\textbf{yolov5m}} \hspace{1em} & SimROD (Ours) w/o Teacher Guidance &  & \checkmark & \checkmark & \checkmark & 71.81 & 18.03\\
\hspace{1em} & SimROD (Ours) w/o Gradual Adaptation & \checkmark & \checkmark &  & \checkmark & 73.45 & 19.67\\
\hspace{1em} & Our full method (SimROD) & \checkmark & \checkmark & \checkmark & \checkmark & 75.40 & 21.62\\
\midrule
\addlinespace[0.3em]
\hspace{1em} & Source &  &  &  &  & 62.84 & 0.00\\
\hspace{1em} & BN-Adapt &  &  & \checkmark &  & 71.83 & 8.99\\
\hspace{1em} & BN-Adapt + DomainMix &  & \checkmark & \checkmark &  & 73.64 & 10.80\\
{\textbf{yolov5x}}\hspace{1em} & SimROD (Ours) w/o Gradual Adaptation & \checkmark & \checkmark &  & \checkmark & 75.58 & 12.74\\
\hspace{1em} & SimROD (Ours) w/o Teacher Guidance &  & \checkmark & \checkmark & \checkmark & 78.16 & 15.32\\
\hspace{1em} & Our full method (SimROD) & \checkmark & \checkmark & \checkmark & \checkmark & 78.48 & 15.64\\
\bottomrule
\end{tabular}
\vspace{-0pt}
\caption{\label{supp:tab:ablation} \small Ablation study on Pascal-C dataset}
\end{table*}

 \section{ Dataset and DomainMix visualizations} \label{supp:visualizations}
\begin{figure*}
\includegraphics[width=2.1\columnwidth]{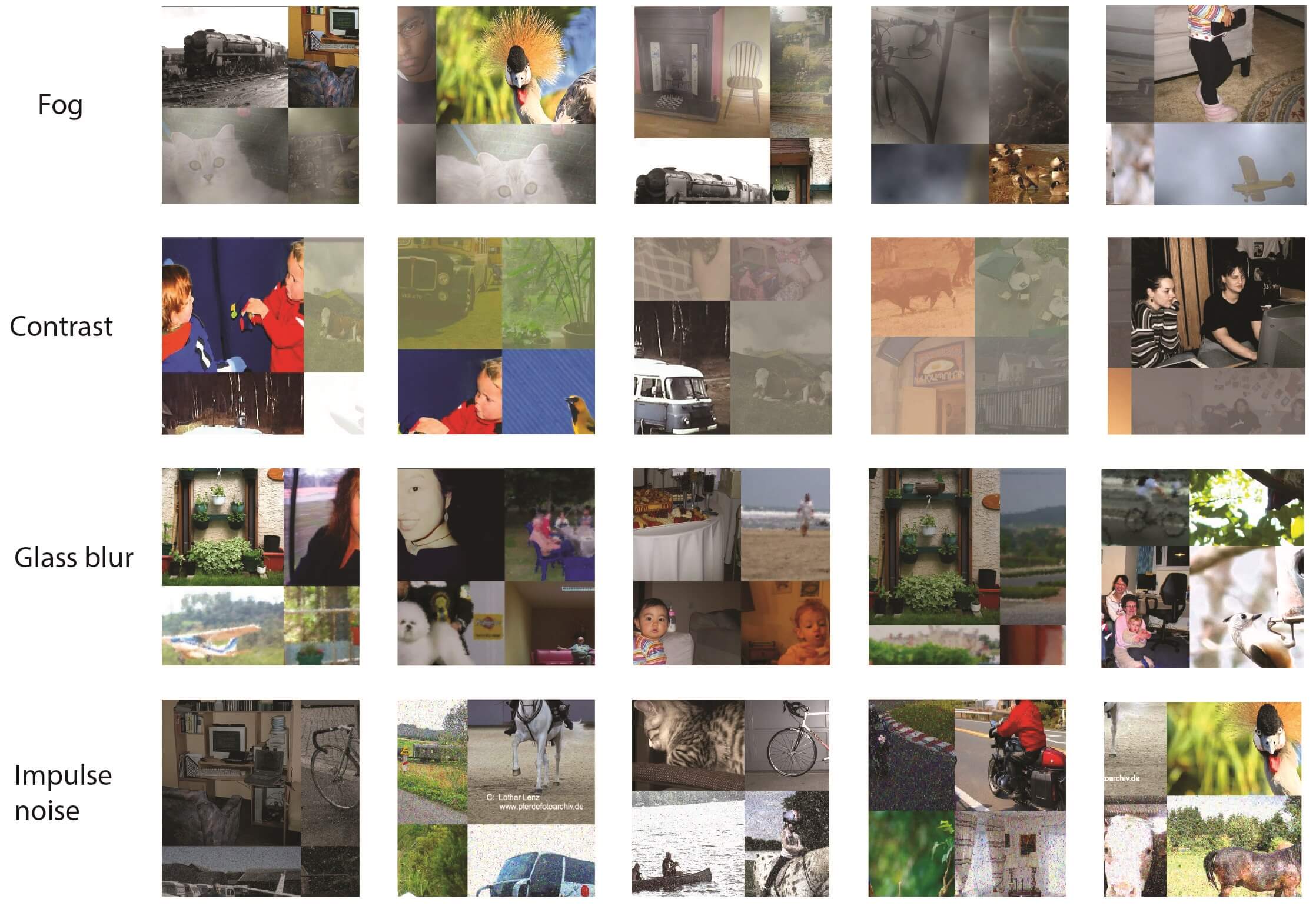}
\vspace{-18pt}
\caption{\small Examples of DomainMix image samples on Pascal-C dataset with various corruption types.}
\label{supp:domainmix-examples-pascal}
\end{figure*}

\begin{figure*}
\includegraphics[width=2.1\columnwidth]{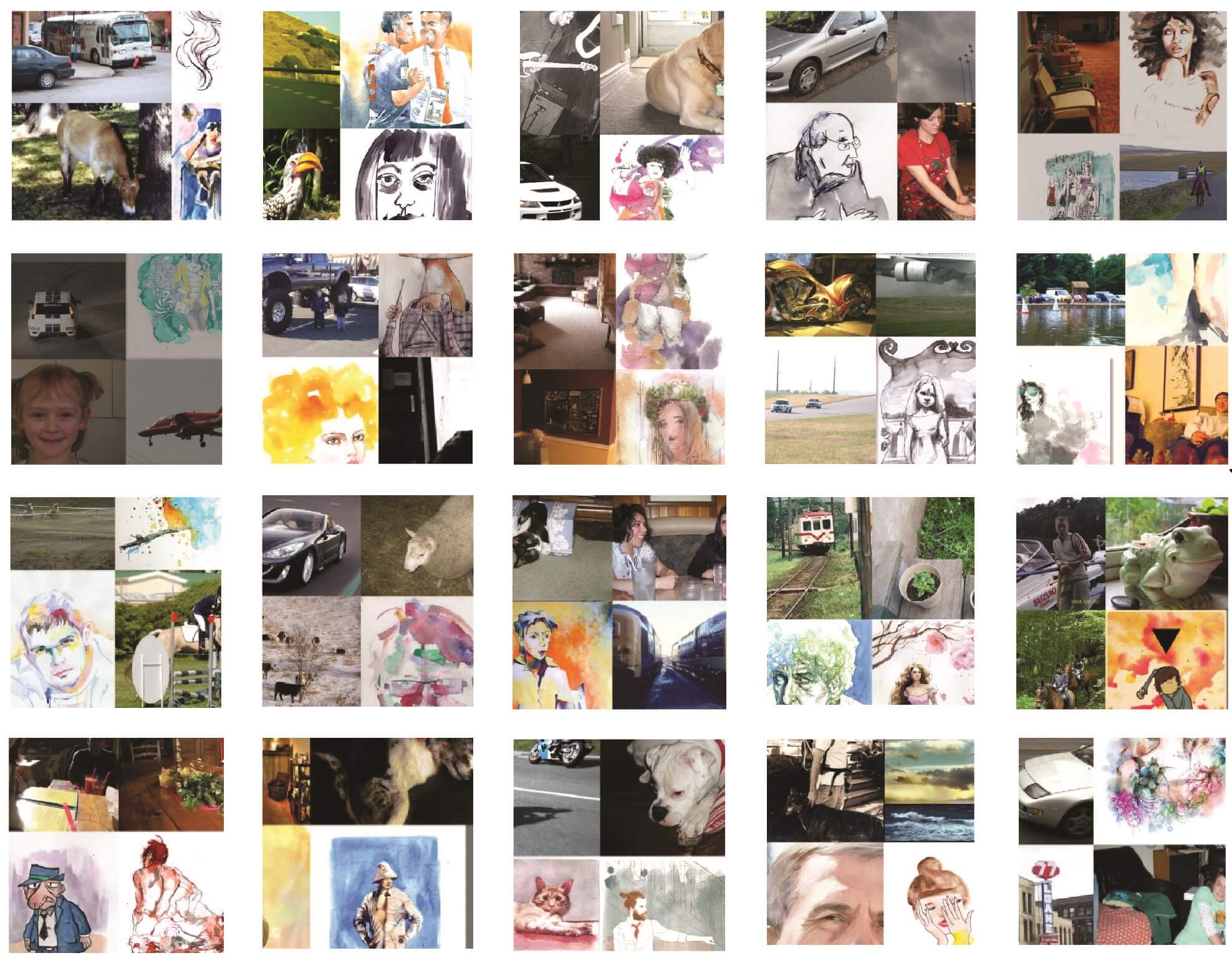}
\vspace{-18pt}
\caption{\small Examples of DomainMix image samples on Watercolor dataset.}
\label{supp:domainmix-examples-watercolor}
\end{figure*}

Fig. \ref{supp:domainmix-examples-pascal} and \ref{supp:domainmix-examples-watercolor} show examples of the domain-mixed images produced by the DomainMix augmentation from different datasets. Note that the images used to form domain-mixed examples, are randomly cropped, and may occupy a different height and width of the final image.

\end{document}